\documentclass[10pt,twocolumn,letterpaper]{article}

\usepackage[]{cvpr}      

\definecolor{cvprblue}{rgb}{0.21,0.49,0.74}
\usepackage[pagebackref,breaklinks,colorlinks,allcolors=cvprblue]{hyperref}
\usepackage{threeparttable}
\usepackage{graphicx}
\usepackage{amsmath}
\usepackage{xcolor,colortbl}
\usepackage{amssymb}
\usepackage{booktabs}
\usepackage{multirow}

\def\paperID{15303} 
\def\confName{CVPR}
\def\confYear{2025}

\title{
Gazing Into Missteps: Leveraging Eye-Gaze for Unsupervised \\Mistake Detection in Egocentric Videos of Skilled Human Activities
}

\author{Michele Mazzamuto$^{1 *}$, 
        Antonino Furnari$^{1 *}$, 
        Yoichi Sato$^{2}$, 
        Giovanni Maria Farinella$^{1}$\\
        \vspace{-0.5cm} 
        \\ 
        $^{1}$University of Catania, Catania, Italy\\
        {\tt\small michele.mazzamuto@phd.unict.it, antonino.furnari@unict.it, giovanni.farinella@unict.it} \\ 
        $^{2}$The University of Tokyo, Tokyo, Japan\\
        {\tt\small ysato@iis.u-tokyo.ac.jp} \\ 
     \small *{Co-first\ authors.}
}

\begin{document}
\maketitle

\begin{abstract}
We address the challenge of unsupervised mistake detection in egocentric video of skilled human activities through the analysis of gaze signals. While traditional methods rely on manually labeled mistakes, our approach does not require mistake annotations, hence overcoming the need of domain-specific labeled data. Based on the observation that eye movements closely follow object manipulation activities, we assess to what extent eye-gaze signals can support mistake detection, proposing to identify deviations in attention patterns measured through a gaze tracker with respect to those estimated by a gaze prediction model. Since predicting gaze in video is characterized by high uncertainty, we propose a novel \textit{gaze completion task}, where eye fixations are predicted from visual observations and partial gaze trajectories, and contribute a novel gaze completion approach which explicitly models correlations between gaze information and local visual tokens. Inconsistencies between predicted and observed gaze trajectories act as an indicator to identify mistakes. Experiments highlight the effectiveness of the proposed approach in different settings, with relative gains up to  $+14\%$, $+11\%$, and $
+5\%$ in EPIC-Tent, HoloAssist and IndustReal respectively, remarkably matching results of supervised approaches without seeing any labels.
We further show that gaze-based analysis is particularly useful in the presence of skilled actions, low action execution confidence, and actions requiring hand-eye coordination and object manipulation skills. Our method is ranked first on the HoloAssist Mistake Detection challenge.
\end{abstract}    

\begin{figure}
    \centering
    \captionsetup{type=figure}
    \includegraphics[width=\linewidth]{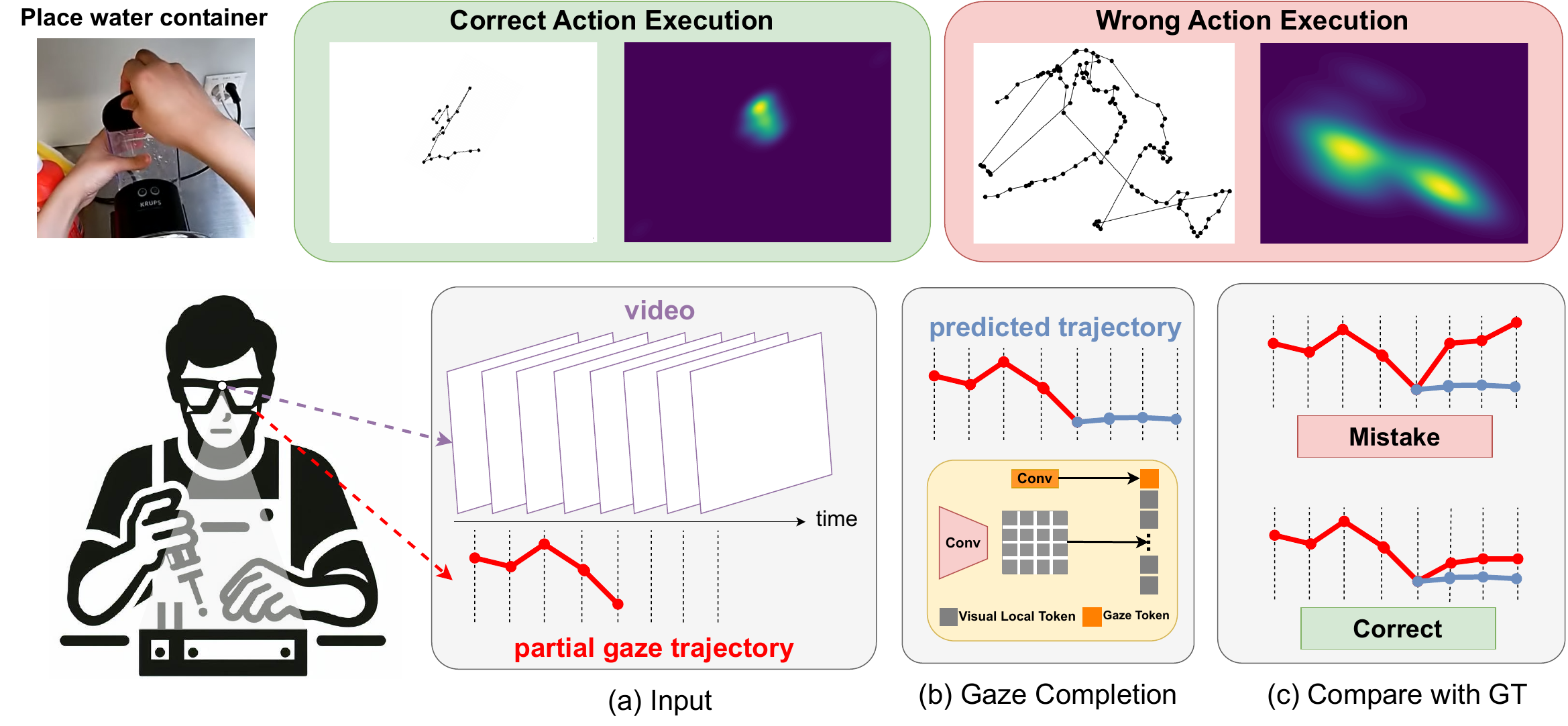}
    \caption{Top: gaze trajectories of a correct and wrong execution of the ``place water container'' action, together with gaze fixation maps averaged across many action instances. Note the higher variability exhibited by wrong executions. Bottom: (a) The proposed unsupervised mistake detection method assumes as input a video with a partial gaze trajectory on the initial part of the video. (b) A gaze completion model predicts a gaze trajectory for the remaining part of the video, conditioned on the input video and the partial trajectory. (c) A mistake is detected if the predicted trajectory is significantly different from the observed one, suggesting a deviation from the expected attention patterns.}
    \label{fig:concept}
\end{figure}

\section{Introduction}
\label{sec:intro}

Smart glasses are gaining more and more popularity, with various existing products capable of supporting the user through Augmented Reality.
In order to provide timely assistance, wearable devices should be able to identify moments in which the user makes mistakes or is confused and requires help~\cite{feng2024struggling}.
If such instances are properly detected, the AI system can proactively 
offer contextual information or suggestions on how to best carry out the task at hand~\cite{HoloAssist2023}. 

Previous works tackled the problem of detecting mistakes from a fully supervised perspective, where mistake instances were labeled in egocentric video and machine learning algorithms were trained to discriminate between video segments of correct action executions and incorrect ones~\cite{sener2022assembly101,HoloAssist2023}. Such a fully supervised approach has two main downsides: 1) It is domain-dependent, hence requiring an accurate characterization of what a mistake is, depending on the context (e.g., a mistake in a kitchen is different from a mistake in the assembly line); 2) It requires to collect and label a sufficient number of mistake instances, which may be difficult to observe and record, involve time consuming procedures, and require expert knowledge.
Another class of methods~\cite{flaborea2024prego,seminara2024differentiable} aim to detect mistakes without relying on mistake annotations, but still requires domain-specific and costly temporal action annotations.
Ideally, a wearable assistant should be able to infer when the behavioral patterns of the user deviate from the norm in order to determine if they need assistance in a scenario-independent setting, i.e., without making specific assumptions on how a mistake is defined and without requiring costly labels.
To overcome the aforementioned limitations, we propose to detect mistakes in egocentric videos of human activity in an unsupervised way, learning from unlabeled video.

Mistakes in task execution, in particular for tasks requiring hand-eye coordination and object manipulation skills, often involve abnormal attention patterns of the camera wearer~\cite{Krafka,Koochaki}. For instance, imagine a user operating a coffee machine without first adding water. As they press the \textit{brew} button, they notice that no coffee is produced and start shifting their attention erratically between the cup, the water tank, and the LED indicator, deviating from the typical attention sequence of “button $\to$ cup $\to$ button.” (See Figure~\ref{fig:concept}(top)). This behavior is well-documented in psychology literature, which shows that gaze patterns are crucial for the execution of even the most repetitive daily activities (e.g., making tea)~\cite{gaze1999}, and that they change in response to task complexity~\cite{Pelz} and mistakes~\cite{Peacock}.

Following these observations, we study how the analysis of eye-gaze fixations can support mistake detection in egocentric video of skilled human activities. We hence propose to learn a model of ``normal'' attention patterns in the form of a gaze predictor producing likely eye gaze trajectories from a video at inference time.
Since gaze prediction can be governed by high uncertainty, depending on the user's goals, we propose a novel ``gaze completion'' task in which a model takes as input a video and a partial gaze trajectory (Figure~\ref{fig:concept}(a)) and is tasked to predict a likely continuation of the partial trajectory (Figure~\ref{fig:concept}(b)).
Gaze completion is tackled with a novel approach based on a Gaze-Frame Correlation module which explicitly models the correlation between gaze information and each local visual token.
We expect videos of correct action executions to represent normal user behavior, and hence to be characterized by predictable gaze patterns, while human behavior will deviate from normality, and  gaze will be unpredictable, when a mistake is made by the user.
We hence signal a mistake by comparing the predicted gaze with the ground truth eye gaze trajectory obtained through a gaze tracker (Figure~\ref{fig:concept}(c)). 

Experiments show the effectiveness of the proposed approach, both alone, or in combination with other techniques, when compared to one-class anomaly detection methods~\cite{TrajREC_stergiou2024holistic,MoCoDAD_Flaborea_2023_ICCV}, and various unsupervised mistake detection baselines, with relative gains up to $+14\%$, $+11\%$, and $+5\%$ in the on EPIC-Tent~\cite{Tent}, HoloAssist~\cite{HoloAssist2023} and IndustReal~\cite{schoonbeek2024industreal} datasets respectively, remarkably matching the results of supervised methods without any labels in one-class settings.
Our analysis also shows that gaze is most effective in the presence of complex actions, low-confidence executions, and actions requiring hand-eye coordination and object-manipulation skills.

In summary, the contributions of this work are as follows:
1) We investigate for the first time the problem of unsupervised mistake detection from egocentric video of human activity and provide an initial benchmark based on three datasets.
2) We define the novel ``gaze completion'' task where models predict gaze trajectories from video and partial gaze inputs, and introduce an approach based on a Gaze-Frame Correlation module;
3) We propose an approach to unsupervised mistake detection leveraging gaze completion to identify instances of unpredicable gaze patterns. 
Experiments analyze under which conditions gaze-based analysis is most useful and show the effectiveness of the approach, in one-class and unsupervised settings. 

{We will publicly release the code and model checkpoints to support future research.}

\section{Related Work}

\noindent
\textbf{Egocentric gaze estimation}
Literature on gaze estimation from egocentric video is rich, with previous works investigating simultaneous gaze prediction and action recognition~\cite{fathi2012learning}, describing gaze prediction approaches incorporating egocentric cues~\cite{li2013learning}, modeling task-dependent attention transition~\cite{Huang2018PredictingGI}, leveraging vanishing point, manipulation point, hand regions~\cite{8658619}, introducing specific architectures~\cite{Naser,lai2022eye}, and proposing datasets to study egocentric gaze estimation and its applications in a variety of scenarios~\cite{Li_2018_ECCV,Ego4D2022CVPR,EPIC-Tent9022634,HoloAssist2023,schoonbeek2024industreal}.
We propose a novel gaze completion task and show its application to the problem of unsupervised mistake detection in egocentric video. Differently from previous works, we define and tackle the novel task of gaze completion, with the aim to reduce the uncertanty associated with gaze prediction.

\noindent
\textbf{Use of gaze in egocentric vision }
While many previous works focused on gaze estimation from video, few works investigated the use of gaze, estimated through a dedicated gaze tracker, as an input to support downstream egocentric vision applications. Specifically, previous investigations focused on discovering object usage~\cite{Damen2014YouDoID}, detecting privacy-sensitive situations~\cite{steil2019privaceye},  finding attended objects~\cite{mazzamuto_gaze}, assisting large language models in classification tasks~\cite{konrad2024gazegpt}, enhancing visual tasks~\cite{zhou2024learningobservergazezeroshotattention,10204584}, improving egocentric human motion prediction~\cite{Zheng2022GIMOGH}, and aiding natural language processing tasks~\cite{NEURIPS2020_460191c7}. We show the effectiveness of gaze in mistake detection. Our method compares gaze trajectories predicted from visual data with gaze estimated through a gaze tracker to identify mistakes when predictions deviates from the ground truth. 

\noindent
\textbf{Mistake Detection in Egocentric Videos}
Mistakes naturally occur in human activities. The ability to automatically detect them from egocentric video can be beneficial for an AR assistant to offer support. 
Identifying mistakes usually entails modeling procedural knowledge~\cite{ding2023mistakeYao,flaborea2024prego,sener2022assembly101}, skill assessment~\cite{gao2014jhu}, action segmentation~\cite{Ghoddoosian_2023_ICCV} or detecting forgotten actions~\cite{Lights_Off}. 
Notably, previous works tackled the task in a supervised fashion, training models to classify an action segment as ``correct'' or ``mistake'' in manually annotated instances~\cite{sener2022assembly101,HoloAssist2023}.
While this approach is feasible in a closed-world scenario, it requires 1) a definition of what a mistake is, depending on the domain (e.g., kitchens vs the assembly line), 2) significant amounts of manually labeled data, which is expensive and requires expert knowledge.
In this work, we tackle an unsupervised mistake detection task, in which models observe unlabeled video at training time and are tasked to detect mistakes from video at test time. Our unsupervised scheme is possible through the analysis of gaze attention patterns, which provide a supervisory signal to create a joint video-gaze model of normal behavior.



\noindent
\textbf{Video Anomaly Detection}
Our research also relates to the problem of Video Anomaly Detection (VAD), which involves recognizing abnormal or anomalous events within videos~\cite{TrajREC_stergiou2024holistic,liu2021hf2vad}.
A line of video anomaly methods are based on one-class classification, in which models are trained on normal videos and aim to identify divergence from the norm at test time~\cite{6751449,Xu2015LearningDR,Sultani_2018_CVPR,Lv2023UnbiasedMI,Feng2021MISTMI,MoCoDAD_Flaborea_2023_ICCV,TrajREC_stergiou2024holistic}.
Notably, anomaly detection in egocentric vision remains under-explored~\cite{Masuda}. 
Similar to video anomaly detection, we aim to detect mistakes by determining video segments which deviate from statistics observed at training time~\cite{MoCoDAD_Flaborea_2023_ICCV}. Differently from previous works in video anomaly detection, we ground our predictions in an egocentric gaze estimation model, which acts as a proxy for modeling normal human behavior, hence effectively achieving mistake prediction detection when anomalous behavior is observed. Moreover, we go beyond the one-class assumption and show that our method can also be used in unsupervised settings where unlabeled correct and mistake examples are included at training time.

\begin{figure*}[t]
  \centering
  \includegraphics[width=0.90\textwidth]{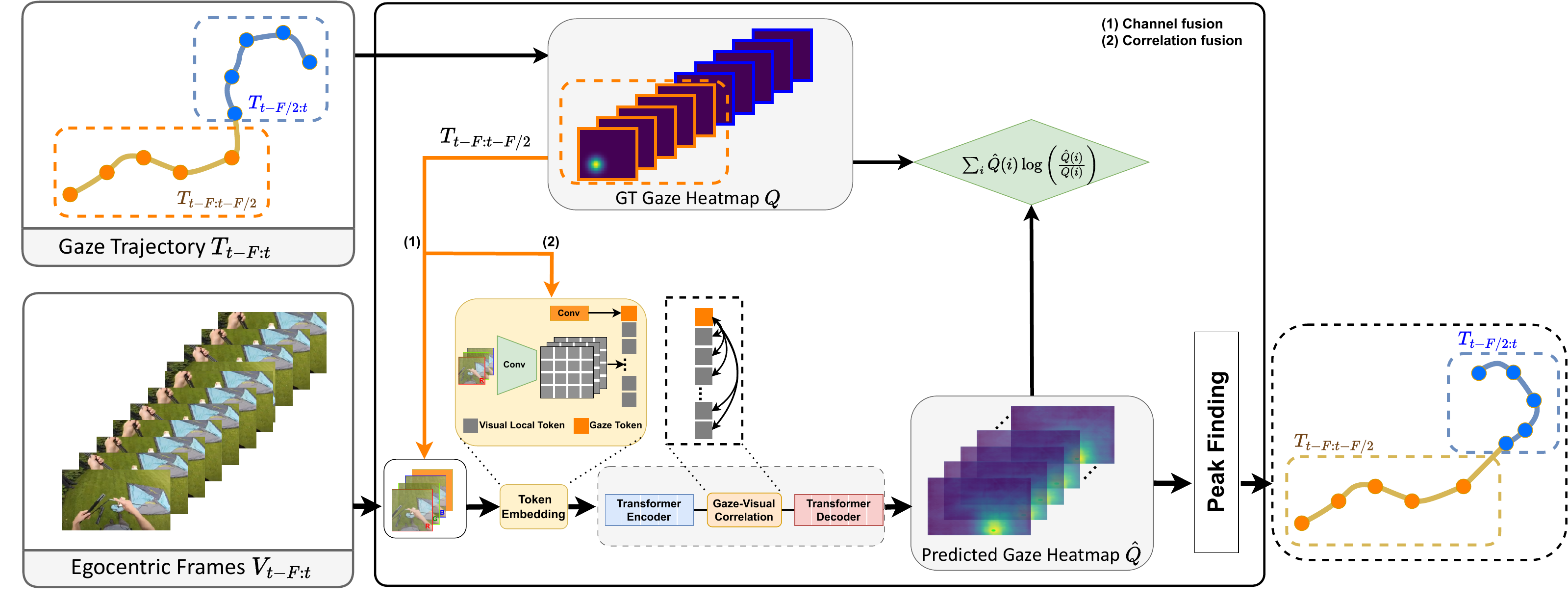}
 \caption{The model takes as input \(F\) RGB frames \(V_{t-F:t}\) and a partial 2D gaze trajectory \(T_{t-F:t-F/2}\) on the first $F/2$ input frames, and outputs a predicted trajectory $\hat{T}_{t-F/2:t}$ from the input video, conditioned on the input trajectory.
 The input trajectory is encoded as a spatio-temporal heatmap $Q$.
 Trajectory and RGB inputs are fused using two strategies, channel fusion, which adds gaze heatmaps as a separate channel (1), and correlation fusion, which uses a dedicated gaze-visual correlation module (2).
 We follow the design of~\cite{lai2022eye} and process our inputs with a transformer encoder-decoder architecture which outputs a predicted gaze heatmap $\hat Q$, supervised via a standard Kullback–Leibler divergence loss using the ground truth unobserved trajectory $T_{t-F/2:t}$.
 The output trajectory $\hat T_{t-F/2:t}$ is recovered from $\hat Q$ using a peak finding operation.}
  \label{fig:approccio_2}
\end{figure*}

\section{Proposed Approach}

\subsection{Mistake Detection Problem Setup}
The mistake detection task consists in highlighting those parts of the video in which the user is making a mistake during the execution of a given activity. In our setup, at each time-step $t$, a model $\Phi$ takes as input a video $V$ observed up to time-step $t$, $V_{1:t}$, and a 2D gaze trajectory $T_{1:t}$ obtained with a gaze tracker, where the $i$-th element of the trajectory $T_i^{(x,y)}$ is a 2D gaze fixation in frame $V_i$. Given this input, the model has to return a score $s_t = \Phi(V_{1:t},T_{1:t})$ indicating whether a mistake is happening at the current time $t$. In this context, high $s_t$ scores indicate the occurrence of a mistake, while low $s_t$ scores indicate a correct action. We can hence see the mistake detection problem as a classification task, in which timestep $t$ is classified as a mistake if $s_t>\theta$, where $\theta$ is a chosen threshold. We follow previous literature on anomaly detection~\cite{TrajREC_stergiou2024holistic,MoCoDAD_Flaborea_2023_ICCV} and evaluate methods in a threshold-independent fashion by reporting the Receiver Operating Characteristics Area Under the Curve (ROC-AUC), where we consider ``mistake'' as the positive class\footnote{A true positive is a mistake correctly classified as a mistake, a true negative is a correct execution correctly classified as a correct execution, a false positive is a correct execution wrongly classified as a mistake, and a false negative is a mistake wrongly classified as a correct execution.}. For completeness, we also report the best $F_1$ score achieved considering the different thresholds, as well as its related precision and recall values.

\subsection{Proposed Approach}
At each timestep $t$, we trim the input video $V_{1:t}$ and gaze trajectory $T_{1:t}$ to the last observed $F$ frames, hence considering $V_{t-F:t}$ and $T_{t-F:t}$ as inputs to our mistake detection method. (Figure~\ref{fig:concept}(a)). Our method relies on two main components: a gaze completion model (Figure~\ref{fig:concept}(b)), and a scoring function (Figure~\ref{fig:concept}(c)).

\noindent
\textbf{Gaze Completion Model}
Figure~\ref{fig:approccio_2} illustrates the proposed gaze completion model.
The model takes as input the video $V_{t-F:t}$ and the first half of the input gaze trajectory $T_{t-F:t-F/2}$ and predicts a gaze trajectory $\hat T_{t-F/2:t}$ aligned to the remaining part of the ground truth trajectory $T_{t-F/2:t}$.
The goal of this model is to predict where the user is looking in the video, conditioned on the initial trajectory. As we show in the experiments, the conditioning allows to reduce the uncertainty on gaze predictions and give a prior into the intention and characteristics of the user. For instance, the model can notice that the user is a novice from the partial input trajectory or get an understanding of the performed activity and adapt its prediction accordingly.
We build on~\cite{lai2022eye} and propose an encoder-decoder transformer-based architecture, including two
approaches to condition gaze prediction on the input partial trajectory: channel fusion and correlation fusion.

\noindent
\textit{Model Overview}
The input gaze trajectory $T_{t-F:t}$ is encoded into a stack of heatmaps $Q$ obtained by centering a Gaussian distribution of standard deviation $\sigma$
around the gaze points.
The first half of the stack $Q_{1:F/2}$, corresponding to the input half trajectory $T_{t-F:t-F/2}$, is forwarded to the two trajectory fusion models (paths (1) and (2) in Figure~\ref{fig:approccio_2}), which inject information on the input trajectory at different semantic levels in the model.
Input frames $V_{t-F:t}$ are processed by a token embedding layer which maps them to visual tokens with a convolution as in~\cite{lai2022eye}.
This module is also responsible for mapping the input heatmaps to a single trajectory token.
Input tokens are then processed by the transformer encoder, by the gaze-visual correlation module and finally by the transformer decoder to output a likely completion of the gaze in the form of heatmaps $\hat Q$, which are supervised with a standard Kullback–Leibler loss.
Specifically, $\hat Q$ contains $F$ heatmaps related to the $F$ input frames $V_{t-F:t}$.
The final trajectory $\hat T$ is obtained by finding the global maxima of the predicted gaze heatmaps.




\noindent
\textit{Channel Fusion} 
The channel fusion module (Figure~\ref{fig:approccio_2}(1)) adds the heatmaps in $Q_{1:F/2}$ to the first $F/2$ frames of the input video, $V_{t-F:t-F/2}$ as an additional channel. The values of this channel are set to zero for the remaining frames. This form of early fusion acts as a \textit{soft conditioning} aiming to include information about the input gaze trajectory in the computation. Note that, in order to incorporate information about the relationships between haze and input frames, the model needs to learn how to compute suitable gaze representations from the additional input channel during training.

\noindent
\textit{Correlation Fusion}
This approach (Figure~\ref{fig:approccio_2}(2)) aims to fuse visual tokens with the gaze trajectory token computed by the token embedding layer. 
Inspired by global-local fusion originally proposed in~\cite{lai2022eye}, this is done in two stages. First, within the transformer encoder, where attention between all visual features and the gaze token is computed. In this stage, correlations between all visual tokens and across visual and gaze tokens are leveraged to obtain a strong representation. Second, within the dedicated gaze-visual correlation module. Here, attention is computed only between the gaze token and the visual tokens, thus learning a dedicated attention mechanism which explicitly enriches the representation of each visual token with gaze information. 
Note that this fusion mechanism operates both at the early (through the encoder) and mid (through the gaze-visual correlation module) levels, thus allowing to leverage low-level (co-occurrences of gaze and visual features) and more semantic (co-occurrences of gaze and semantic visual concepts) information.

\noindent
\textbf{Scoring Function}
Our approach predicts the mistake confidence score $s_t$ by comparing the predicted trajectory $\hat T_{t-F/2:t}$ with the ground truth one $T_{t-F/2:t}$ which is obtained by the gaze tracker of the wearable device. We explore four different ways to compare the two trajectories: Euclidean distance, Dynamic Time Warping (DTW), Heatmap, and Entropy.


\noindent
\textit{Euclidean Distance}
This method consists in accumulating the Euclidean distances computed between corresponding points in each trajectory as the score $s_t$:
\begin{equation}
s_t = \sum_{i=F/2+1}^{F} \|T_{t-F+i} - \hat T_{t-F+i}\|.
\end{equation}


\noindent
\textit{Dynamic Time Warping}
This scoring function uses Dynamic Time Warping (DTW)\cite{Sakoe1978DynamicPA} to measure the distance between trajectories: 

\begin{equation}
    s_t = \texttt{DTW}(T, \hat{T})
\end{equation}
Where $\mathbf{\texttt{DTW}}$ returns the cost of aligning $T$ to $\hat{T}$ according to the DTW algorithm\footnote{We used this implementation: \url{https://pypi.org/project/fastdtw/}.}.

\noindent
\textit{Heatmap}
Differently from previous functions, this approach explicitly considers the probability values predicted by the model at each location.
Specifically, we evaluate the likelihood of a ground truth eye fixation $T_i$ obtained by the device under the predicted heatmap $\hat Q_i$, which can be computed as:

\begin{equation}
    P(T_i|\hat Q) = \hat Q_i(T_i^x, T_i^y)
\end{equation}

\noindent
where $T_i^x$ and $T_i^y$ are the coordinates of the trajectory point $T_i = [T_i^x, T_i^y]$. The score associated to the predicted trajectory $\hat T$ is computed as the sum of the likelihoods of each trajectory point, considering the predicted heatmap $\hat Q$:

\begin{equation}
    s_t = \sum_{i=F/2+1}^F P(T_{t-F+i}|\hat Q).
\end{equation}

\noindent
\textit{Entropy}
This is the only method which does not require ground truth gaze for computation. We consider this measure as a way to check whether mistakes are systematically characterized by uncertain gaze predictions. 
 In this case, the score $s_t$ is set as the mean entropy of all predicted heatmaps for a given trajectory \( \hat{T} \). The entropy \( H \) of a single heatmap \( \hat{Q} \) is given by:
\begin{equation}
s_t = -\frac{1}{F/2}\sum_{i={ F/2}+1}^{F} \sum_{x,y} \hat Q_{t-F+i}^{(x,y)} \log_2(\hat Q_{t-F+i}^{(x,y)})
\end{equation}
$Q_j^{(x,y)}$ is the value at coordinates $(x,y)$ of heatmap $Q_j$.




\section{Experiments and Results}

\subsection{Datasets and Implementation Details}
We perform our experiments on three popular datasets.

\noindent
\textbf{EPIC-Tent}~\cite{EPIC-Tent9022634} includes $7$ hours of egocentric video of $29$ subjects wearing a head-mounded GoPro and an SMI eye tracker while assembling a camping tent. The dataset includes egocentric video, gaze and labels indicating video segments in which users make mistakes. Subjects also rated their level of confidence in action execution in each clip in the videos.
EPIC-Tent contains $151,689$ mistake frames and $384,558$ frames of correct executions, hence with a 28:72 ratio between correct and mistake frames. Since no official train-test split is available, we randomly split videos in training, validation and test sets roughly following a 60:15:25 ratio, obtaining $86,099$, $27,613$, and $37,977$ mistake frames in the training, validation, and test sets respectively.
Since the dataset contains a single video per subject, there is no subject overlap between the three sets.
 

\noindent
\textbf{IndustReal}~\cite{schoonbeek2024industreal} is designed for studying procedural tasks in industrial-like environments and consists of distinct training, validation, and test sets. The training set comprises $78,902$ frames, with $95.68\%$ frames labeled as correct and $4.32\%$ labeled as mistakes. The validation set includes $38,036$ frames, with $95.18\%$ correct and $4.82\%$ mistaken frames. The test set contains $90,105$ frames, with $92.53\%$ correct and $7.47\%$ mistaken frames.

\noindent
\textbf{HoloAssist}~\cite{HoloAssist2023} focuses on a variety of scenarios in which users perform tasks with the assistance of an expert. The training set comprises $11,614,033$ frames, with $94\%$ frames labeled as correct and $6\%$ labeled as mistakes. The test set contains $1,699,562$ frames, with $95\%$ correct and $5\%$ mistake frames.

\noindent
\textbf{Implementation Details}
We process input frames with a stride of $1$ and set the batch size to $4$ clips of $8$ frames each. Weight decay is set to $0.07$ to prevent overfitting. See the supplementary material for more details.


\subsection{Supervision Levels and Compared Approaches}
We compare the proposed approach to methods belonging to three different supervision levels: fully supervised, one-class classification, and unsupervised. All baselines described below are compared with a random baseline which assigns a random score to each input clip.

\noindent
\textbf{Fully Supervised Methods} are trained assuming the availability of mistake labels for all image frames. We consider this class of methods to provide an upper-bound to performance when assessing one-class and unsupervised methods. We consider two approaches in this class: a TimeSformer~\cite{timesformer_bertasius2021spacetime} action recognition model which classifies the input video clips without access to any temporal context from actions executed before or after the current one, and a C2F~\cite{singhania2023c2f} temporal action segmentation model operating on DINOv2~\cite{oquab2023dinov2} features, which naturally performs action segmentation taking into account the temporal context in which an action (or mistake) is executed.


\noindent
\textbf{One-Class Classification Methods} are trained only on \textit{videos of correct executions}, following the standard setup of anomaly detection~\cite{MoCoDAD_Flaborea_2023_ICCV,TrajREC_stergiou2024holistic}. In this context, we assume that the data is verified by an expert for correctness before being used for training. Note that this check does not require marking the temporal occurrence of mistakes, but only discarding any video which contains mistakes.
For this class, we compare our method with respect to TrajREC~\cite{TrajREC_stergiou2024holistic} and MoCoDAD~\cite{MoCoDAD_Flaborea_2023_ICCV}, two popular approaches for video anomaly detection based on the processing of human skeletal data. Since full human skeletons are not visible in egocentric videos, we replace skeletal data with hand joint keypoints\footnote{We use ground truth hand keypoints in HoloAssist and IndustReal, while we extract keypoints with \url{https://github.com/open-mmlab/mmpose} in EPIC-Tent.}. We also adapt TrajREC and MoCoDAD to take a single gaze point instead of, or in addition to, the hand keypoints to assess the ability of such methods to leverage eye-gaze information\footnote{See supplementary material for more details.}.
We compare these models to an instantiation of the proposed approach in which the gaze completion model is trained only on correct executions, hence effectively replicating a one-class scheme. 
We also compare with respect to a baseline which replaces the proposed gaze completion module with a simple gaze prediction component based on GLC~\cite{lai2022eye}.
Following the one-class setup, we train the GLC method of~\cite{lai2022eye} on correct executions only and compare the predicted and ground truth gaze using the considered scoring functions.

\noindent
\textbf{Unsupervised Methods} assume no knowledge of which examples are correct executions and which are mistakes. Hence, models are trained on a natural mix of correct and incorrect action executions. This is the least constrained case in which the collected data is not verified by an expert prior to training.
We compare our model with TrajREC and MoCoDAD adapted as discussed above and with the gaze-prediction baseline GLC~\cite{lai2022eye}.


\begin{table}[t]
    \centering
        \centering

        \resizebox{0.99\linewidth}{!}{
\begin{tabular}{rllcccccc}
\hline
& \textbf{Scoring}  & \textbf{Fusion}&  \textbf{F1} & \textbf{Precision} & \textbf{Recall} & \textbf{AUC} \\
\hline
1&Random& // & 0.36 & 0.29 & 0.42 & 0.51 \\
\hline
2&Entropy&	//&	0.41&	0.27&	0.62&	0.51\\
3&Euclidean&	//&	0.42&	0.29&	0.60&	0.55\\
4&DTW&	//&	0.44&	0.31&	0.68&	0.56\\
5&Heatmap&	//&	\underline{0.45}&	\underline{0.32}&	\underline{0.70}&	\underline{0.57}\\
\hline
6&Heatmap&	CH&	0.45&	0.32&	0.74&	0.63\\
7&Heatmap&	CORR&	\underline{0.50}&	\underline{0.36}&	\underline{0.82}&	\underline{0.65}\\
\hline
8&Heatmap&	 CH + CORR &	\textbf{\underline{0.51}}&	\textbf{\underline{0.36}}&	\textbf{\underline{0.85}}&	\textbf{\underline{0.69}}\\
\hline
\end{tabular}}  
\caption{Ablation of various scoring functions and fusion strategies on EPIC-Tent in the unsupervised setting. Best results per-block are \underline{underlined}, while best global results are \textbf{in bold}. CH: channel fusion, CORR: correlation fusion.}
 \label{tab:ablation}
\end{table}



\subsection{Performance of Proposed Model and Ablations}
Table~\ref{tab:ablation} reports the performance of the proposed approach in unsupervised settings, evaluating the considered scoring functions and gaze-video fusion strategies in the unsupervised mistake detection settings on EPIC-Tent. 
Rows 2-5 compare scoring functions when both fusion strategies are turned off and the model is not conditioned on previous trajectories.
The entropy scoring function achieved an AUC of $0.51$ and an F1 score of $0.41$ (row 2), only marginally above the random baseline (F1 of $0.36$), suggesting that high entropy in the predictions marginally correlates with the presence of a mistake. 
Alternative scoring functions improved the results, yielding an AUC of $0.55$ and $0.56$ when using Euclidean distance and DTW scoring functions, respectively (rows 3-4). The heatmap-based scoring function produced the best results, with an AUC of $0.57$ and an F1 score of $0.45$, significantly above random level (compare with the F1 score of $0.36$ in row 1). 
The advantage of the heatmap scoring function is likely due to the better exploitation of the probability values computed by the gaze prediction model, as compared to other scoring functions.
Hence, we adopted the heatmap-based scoring as our primary method in following comparisons.

Rows 6-7 compare approaches using one of the two fusion strategies. While both fusion strategies improve results (compare rows 6-7 with 5), the proposed correlation strategy (CORR) systematically outperforms channel fusion, obtaining an AUC score of $0.65$ and an F1 score of $0.50$ and doubling the recall of the random baseline ($0.82$ vs $0.42$) with better precision ($0.36$ vs $0.29$).
Combining the two fusion strategies (row 8) leads to an AUC of $0.69$ and an F1 score of $0.51$ (\textit{+0.12} and \textit{+0.06} compared to the standard heatmap method - row 5). This configuration is the one referred to as ``ours'' in future comparisons.\footnote{See supplementary material for additional ablations.} 

\begin{table}[t]
    \centering
    
    \begin{threeparttable}
        \resizebox{0.99\linewidth}{!}{
            \begin{tabular}{lcccccc}
                \hline
                \textbf{Method} & \textbf{Sup. Level} & \textbf{F1} & \textbf{Precision} & \textbf{Recall} & \textbf{AUC} \\
                \hline
                Random & // & 0.36 & 0.29 & 0.42 & 0.51 \\
                 TimeSformer~\cite{timesformer_bertasius2021spacetime} & Fully Supervised & \underline{0.49} & \underline{0.35} & \underline{0.80} & \underline{0.67} \\
                 C2F~\cite{singhania2023c2f} & Fully Supervised & \textbf{0.58} & \textbf{0.44} & \textbf{0.85} & \textbf{0.72} \\
                \hline \hline
                TrajREC (G)~\cite{TrajREC_stergiou2024holistic} & One-Class & 0.40 & 0.26 & \underline{0.88} & 0.51 \\
                MoCoDAD (G)~\cite{MoCoDAD_Flaborea_2023_ICCV} & One-Class & 0.43 & 0.27 & \textbf{0.91} & 0.50 \\
                 TrajREC (H)~\cite{TrajREC_stergiou2024holistic} & One-Class & 0.44 & 0.31 & 0.76 & 0.55 \\
                 MoCoDAD (H)~\cite{MoCoDAD_Flaborea_2023_ICCV} & One-Class & 0.46 & 0.33 & 0.79 & 0.60 \\
                 TrajREC (H+G)~\cite{TrajREC_stergiou2024holistic} & One-Class & 0.42 & 0.29 & 0.75 & 0.53 \\
                 MoCoDAD (H+G)~\cite{MoCoDAD_Flaborea_2023_ICCV} & One-Class & 0.43 & 0.30 & 0.77 & 0.56 \\
                 TrajREC (H+G)*~\cite{TrajREC_stergiou2024holistic} & One-Class & 0.47 & 0.34 & 0.77 & 0.63 \\
                 MoCoDAD (H+G)*~\cite{MoCoDAD_Flaborea_2023_ICCV} & One-Class & 0.49 & 0.35 & 0.81 & 0.65 \\
                GLC~\cite{lai2022eye} & One-Class & 0.46 & \underline{0.37} & 0.62 & 0.66 \\
                \hline
                Ours & One-Class & \underline{0.52} & \underline{0.37} & {0.85} & \underline{0.69} \\
                 Ours + MoCoDAD (H)* & One-Class & \textbf{0.54} & \textbf{0.41} & {0.86} & \textbf{0.72} \\
                
                \hline \hline
                TrajREC (G)~\cite{TrajREC_stergiou2024holistic} & Unsupervised & {0.27} & {0.16} & \textbf{0.94} & {0.50} \\
                 MoCoDAD (G)~\cite{MoCoDAD_Flaborea_2023_ICCV} & Unsupervised & {0.33} & {0.21} & \underline{0.88} & {0.51} \\ TrajREC (H)~\cite{TrajREC_stergiou2024holistic} & Unsupervised & {0.40} & {0.27} & {0.79} & {0.58} \\
                  MoCoDAD (H)~\cite{MoCoDAD_Flaborea_2023_ICCV} & Unsupervised & {0.41} & {0.27} & {0.86} & {0.60} \\
                  MoCoDAD (H+G)*~\cite{MoCoDAD_Flaborea_2023_ICCV} & Unsupervised & 0.41 & 0.27 & \underline{0.88} & 0.60 \\
                GLC~\cite{lai2022eye} & Unsupervised & 0.44 & 0.33 & 0.70 & 0.61 \\
                \hline
                Ours & Unsupervised & \underline{0.51} & \underline{0.36} & {0.85} & \underline{0.69} \\
                Ours + MoCoDAD (H)* &Unsupervised&\textbf{0.52}&	\textbf{0.37}&	\underline{0.88}&	\textbf{0.70} \\
                \hline \hline
            \end{tabular}
        }
        \begin{tablenotes}
            \item[*] \footnotesize Late fusion
        \end{tablenotes}
    \end{threeparttable}
    \caption{Mistake detection results on EPIC-Tent. Best results are \textbf{in bold}, second best results are \underline{underlined}.}
    \label{tab:GLC_EPIC}
\end{table}

\subsection{Comparison with the state of the art}
\textbf{EPIC-Tent} Table~\ref{tab:GLC_EPIC} compares the proposed mistake detection approach with competitors on EPIC-Tent, according to the three considered levels of supervision.
C2F outperforms TimeSformer in all evaluated metrics, particularly in the F1 score ($0.58$ of C2F vs $0.49$ of TimeSformer and $0.36$ of the random baseline), suggesting that C2F is more adept at capturing temporal reasoning, which is crucial for identifying mistakes in dynamic activities. However, the reliance on fully labeled datasets poses limitations for both methods. One-class methods for anomaly detection adapted to take only gaze as input, namely, TrajREC (G) and MoCoDAD (G), show minor improvements over the random baseline in terms of AUC score ($0.50 - 0.51$ vs $0.51$) and only small improvements in F1 score ($0.40 - 0.43$ vs $0.36$). High recall values, paired with low precision, suggest that these methods tend to classify most clips as mistakes. Incorporating hand skeleton data instead of gaze, namely, TrajREC (H) and MoCoDAD (H), leads to slight improvements, as evidenced by F1 scores of $0.44$ and $0.45$, and AUC values of $0.55$ and $0.60$. Combining (H) and (G) models through late fusion (denoted with $^*$) improves the results, with MoCoDAD (H+G)$^*$ achieving an F1 score of 0.49 and an AUC of 0.65, suggesting that the signals captured by gaze-based and hand-based analyses are complementary. The one-class approach based on GLC~\cite{lai2022eye} gaze prediction shows improved results compared to previous one-class methods, being only slightly less effective than the late fused method, despite not analyzing any hand-based information. This highlights the value of leveraging gaze analysis for mistake detection.
```

Finally, the proposed method based on gaze completion obtains the best results, yielding an AUC of $0.69$, an F1-score of $0.52$, a precision of $0.37$, and a recall of $0.85$, which amount to relative improvements of $+5/13\%$\footnote{We compute the relative improvement of $b$ with respect to $a$ as $\frac{b-a}{b}$} with respect to the best approach GLC and $+35/44\%$ with respect to the random baseline. Late-fusing our approach with MoCoDAD (H) achieves enhanced results with an AUC of $0.72$ and an F1 score of $0.54$, improving over compared approaches, suggesting that gaze analysis can further benefit from integration with approaches based on different cues. It is worth noting that our best method remarkably achieves the same AUC score of $0.72$ as the best supervised approach and a comparable $F1$ score ($0.54$ vs $0.58$) without access to labels during training. We compare unsupervised approaches in the bottom part of Table~\ref{tab:GLC_EPIC}. 
Similarly to the one-class case, TrajREC (H) and MoCoDAD (H) slightly improve over the random baseline (e.g., $0.40$ and $0.41$ vs $0.36$ of the random baseline in F1 score). GLC outperforms these former two methods obtaining an F1 score of $0.44$ and an AUC score of $0.61$, which are lower than the scores of $0.46$ and $0.66$ obtained in the one-class setting. In these settings, the proposed method achieves an F1 score of $0.51$, which is comparable to the score obtained in one-class settings $0.52$ with similar and recall values and the same AUC of $0.69$, despite the unsupervised setting being more challenging.
Late fusion with MoCoDAD (H) brings some additional improvements, with an F1 score of 0.52 and AUC of 0.70.

\begin{table}[t]
    \centering
        \centering
       
        \begin{threeparttable}
        \resizebox{0.99\linewidth}{!}{
            \begin{tabular}{lcccccc}
                \hline
                \textbf{Method}&\textbf{Sup. Level} &\textbf{F1} & \textbf{Precision} & \textbf{Recall} & \textbf{AUC} \\
                \hline
                Random&//&0.04&	0.02&	\underline{0.39}&	0.50 \\
                 TimeSformer~\cite{timesformer_bertasius2021spacetime}&Fully Supervised&\underline{0.21}&	\underline{0.35}&	0.13&	\underline{0.58} \\
                 C2F~\cite{singhania2023c2f} &Fully Supervised&\textbf{0.38}&	\textbf{0.37}&	\textbf{0.40}&	\textbf{0.65} \\
                \hline \hline
                TrajREC (G)~\cite{TrajREC_stergiou2024holistic} &One-Class&0.09 & 0.04 & \textbf{0.96} & 0.50 \\
                MoCoDAD (G)~\cite{MoCoDAD_Flaborea_2023_ICCV} &One-Class&0.11 & 0.06 & \underline{0.94} & 0.51 \\
                 TrajREC (H)~\cite{TrajREC_stergiou2024holistic} &One-Class&0.19 & 0.11 & 0.72 & 0.56 \\
                 MoCoDAD (H)~\cite{MoCoDAD_Flaborea_2023_ICCV} &One-Class&0.17 & 0.10 & 0.71 & 0.55 \\
                 TrajREC (H+G)~\cite{TrajREC_stergiou2024holistic} &One-Class&0.13 & 0.07 & 0.68 & 0.52 \\
                MoCoDAD (H+G)~\cite{MoCoDAD_Flaborea_2023_ICCV} &One-Class&0.14 & 0.08 & 0.62& 0.52 \\
                TrajREC (H+G)*~\cite{TrajREC_stergiou2024holistic} &One-Class&0.20 & 0.12 & 0.71 & 0.56 \\
                 MoCoDAD (H+G)*~\cite{MoCoDAD_Flaborea_2023_ICCV} &One-Class&0.21 & 0.12 & 0.75 & 0.57 \\
                GLC~\cite{lai2022eye} &One-Class&0.19&	0.11&	0.56&	0.60\\
                \hline
                Ours &One-Class&\underline{0.22}&	\underline{0.14}&	{0.59}&	\underline{0.61} \\
                 Ours + MoCoDAD (H)* &One-Class&\textbf{0.26}&	\textbf{0.16}&	{0.73}&	\textbf{0.63} \\
                \hline \hline

                TrajREC (G)~\cite{TrajREC_stergiou2024holistic} & Unsupervised & {0.05} & {0.03} & \textbf{0.92} & {0.50} \\
                  MoCoDAD (G)~\cite{MoCoDAD_Flaborea_2023_ICCV} & Unsupervised & {0.07} & {0.04} & \textbf{0.92} & {0.50} \\
                 TrajREC (H)~\cite{TrajREC_stergiou2024holistic} & Unsupervised & {0.11} & {0.07} & {0.32} & {0.56} \\
                 MoCoDAD (H)~\cite{MoCoDAD_Flaborea_2023_ICCV} & Unsupervised & {0.14} & {0.10} & {0.25} & {0.55} \\
                 MoCoDAD (H+G)*~\cite{MoCoDAD_Flaborea_2023_ICCV} & Unsupervised & 0.15 & 0.11 & 0.25 & 0.56 \\
                GLC~\cite{lai2022eye} &Unsupervised&0.10&	0.06&0.34&	0.54\\
                \hline
                Ours &Unsupervised&\underline{0.18}&	\underline{0.12}&	\underline{0.40}&	\underline{0.59} \\
                Ours + MoCoDAD (H)* &Unsupervised&\textbf{0.21}&	\textbf{0.15}&	\underline{0.40}&	\textbf{0.60} \\
                \hline \hline
            \end{tabular}
        }
         \begin{tablenotes}
             \item[*]\footnotesize Late fusion 
         \end{tablenotes}
        \end{threeparttable}
         \caption{Mistake detection result on HoloAssist.}
        \label{tab:GLC_HoloAssist}
\end{table}

Table \ref{tab:GLC_HoloAssist} compares the proposed method with competitors on HoloAssist, which presents a more varied and expansive context than EPIC-Tent, making mistake detection more challenging. The random baseline achieves an F1 score of only $0.04$. One-class methods like TrajREC and MoCoDAD improve over the baseline but tend to classify most clips as mistakes, with high recall values ($0.96$ and $0.94$). Hand keypoint-based methods show improvements, with TrajREC (H) slightly outperforming MoCoDAD (H). GLC demonstrates better AUC and F1 scores, with more balanced precision and recall metrics. Our approach achieves an AUC of $0.61$ and an F1 score of $0.22$ in one-class settings, with the best results from late fusion with MoCoDAD, yielding an F1 score of $0.26$ and an AUC of $0.63$. In the unsupervised scenario, TrajREC (G) and MoCoDAD (G) show limited effectiveness, while the (H) approaches perform slightly better. Our method achieves an AUC of $0.59$ and an F1 score of $0.18$, showing robustness across evaluation settings and relative improvements over GLC, with gains of $+9\%$ and $+80\%$, respectively. Combining with MoCoDAD (H) further enhances performance.

Results on IndustReal, shown in Table \ref{tab:GLC_IndustReal}, confirm the trends observed in HoloAssist. TrajREC and MoCoDAD bring small improvements over the random baseline in H, G, and H+G configurations. Our method outperforms competitors with improvements over GLC of $+5\%$ in AUC and $+14\%$ in F1 in one-class settings, and $+6\%$ in AUC in unsupervised settings, while late fusion with MoCoDAD (H) does not improve performance due to MoCoDAD’s reduced effectiveness in this scenario.

\begin{table}[t]
    \centering
        \centering
        
        \begin{threeparttable}
\resizebox{0.99\linewidth}{!}{
    \begin{tabular}{lcccccc}
        \hline
        \textbf{Method} & \textbf{Sup. Level} & \textbf{F1} & \textbf{Precision} & \textbf{Recall} & \textbf{AUC} \\
        \hline
        Random & // & 0.12 & 0.06 & \textbf{0.62} & 0.51  \\
         TimeSformer~\cite{timesformer_bertasius2021spacetime}&Fully Supervised&\underline{0.20}&	\underline{0.12}&	\underline{0.35}&	\underline{0.58} \\
        C2F~\cite{singhania2023c2f}&Fully Supervised&\textbf{0.31}&	\textbf{0.29}&	0.31&	\textbf{0.67} \\
        \hline \hline
        TrajRE(G)~\cite{TrajREC_stergiou2024holistic} & One-Class & 0.17 & 0.09 & \underline{0.90} & 0.53 \\
        MoCoDAD(G)~\cite{MoCoDAD_Flaborea_2023_ICCV} & One-Class & 0.18 & 0.10 & \textbf{0.91} & 0.55 \\
         TrajREC(H)~\cite{TrajREC_stergiou2024holistic} &One-Class&0.21 & 0.12 & 0.88 & 0.57 \\
         MoCoDAD(H)~\cite{MoCoDAD_Flaborea_2023_ICCV} &One-Class&0.22 & 0.13 & 0.81 & 0.60 \\
        
         TrajREC(H+G)~\cite{TrajREC_stergiou2024holistic} &One-Class&0.18 & 0.10 & 0.86 & 0.55 \\
 MoCoDAD(H+G)~\cite{MoCoDAD_Flaborea_2023_ICCV} &One-Class&0.19 & 0.11 & 0.79 & 0.58 \\

TrajREC(H+G)*~\cite{TrajREC_stergiou2024holistic} &One-Class&0.21 & 0.12 & 0.88 & 0.58\\
                 MoCoDAD(H+G)*~\cite{MoCoDAD_Flaborea_2023_ICCV} &One-Class&0.22 & 0.13 & 0.82 & 0.61 \\
        GLC~\cite{lai2022eye} &One-Class& 0.21 & 0.15 & 0.33 & 0.60 \\
        \hline
        Ours & One-Class & \underline{0.24} & \textbf{0.18} & 0.35 & \underline{0.63} \\
         Ours + MoCoDAD (H)* &One-Class&\textbf{0.26}&	\underline{0.17}&	{0.60}&	\textbf{0.65} \\
        \hline \hline
        TrajREC (G)~\cite{TrajREC_stergiou2024holistic} & Unsupervised & {0.11} & {0.06} & \textbf{0.92} & {0.51} \\
                 MoCoDAD (G)~\cite{MoCoDAD_Flaborea_2023_ICCV} & Unsupervised & {0.11} & {0.06} & \textbf{0.92} & {0.51} \\
          TrajREC (H)~\cite{TrajREC_stergiou2024holistic} & Unsupervised & {0.15} & {0.11} & {0.28} & {0.55} \\
                  MoCoDAD (H)~\cite{MoCoDAD_Flaborea_2023_ICCV} & Unsupervised & {0.16} & {0.12} & {0.29} & {0.57} \\
        MoCoDAD (H+G)*~\cite{MoCoDAD_Flaborea_2023_ICCV} & Unsupervised & 0.17 & 0.12 & 0.30 & 0.57 \\
        GLC~\cite{lai2022eye} &Unsupervised& \textbf{0.21} & \underline{0.15} & \underline{0.33} & 0.58 \\
        \hline
        Ours & Unsupervised & \textbf{0.21} & \textbf{0.16} & \underline{0.33} & \textbf{0.62} \\
        Ours + MoCoDAD (H)* &Unsupervised&\underline{0.20}&	\underline{0.15}&	{0.32}&	\underline{0.61} \\
        \hline \hline
    \end{tabular}
}
           \begin{tablenotes}
             \item[*]\footnotesize Late fusion 
         \end{tablenotes}
        \end{threeparttable}
        \caption{Mistake detection result on IndustReal.}
        \label{tab:GLC_IndustReal}
\end{table}

\subsection{Contribution of gaze across scenarios}
In this section, we analyze the performance of our method with respect to scenarios in order to assess under which conditions gaze analysis is more or less predictive of mistakes.

\begin{figure}[t]
    \centering
    \includegraphics[width=0.90\linewidth]{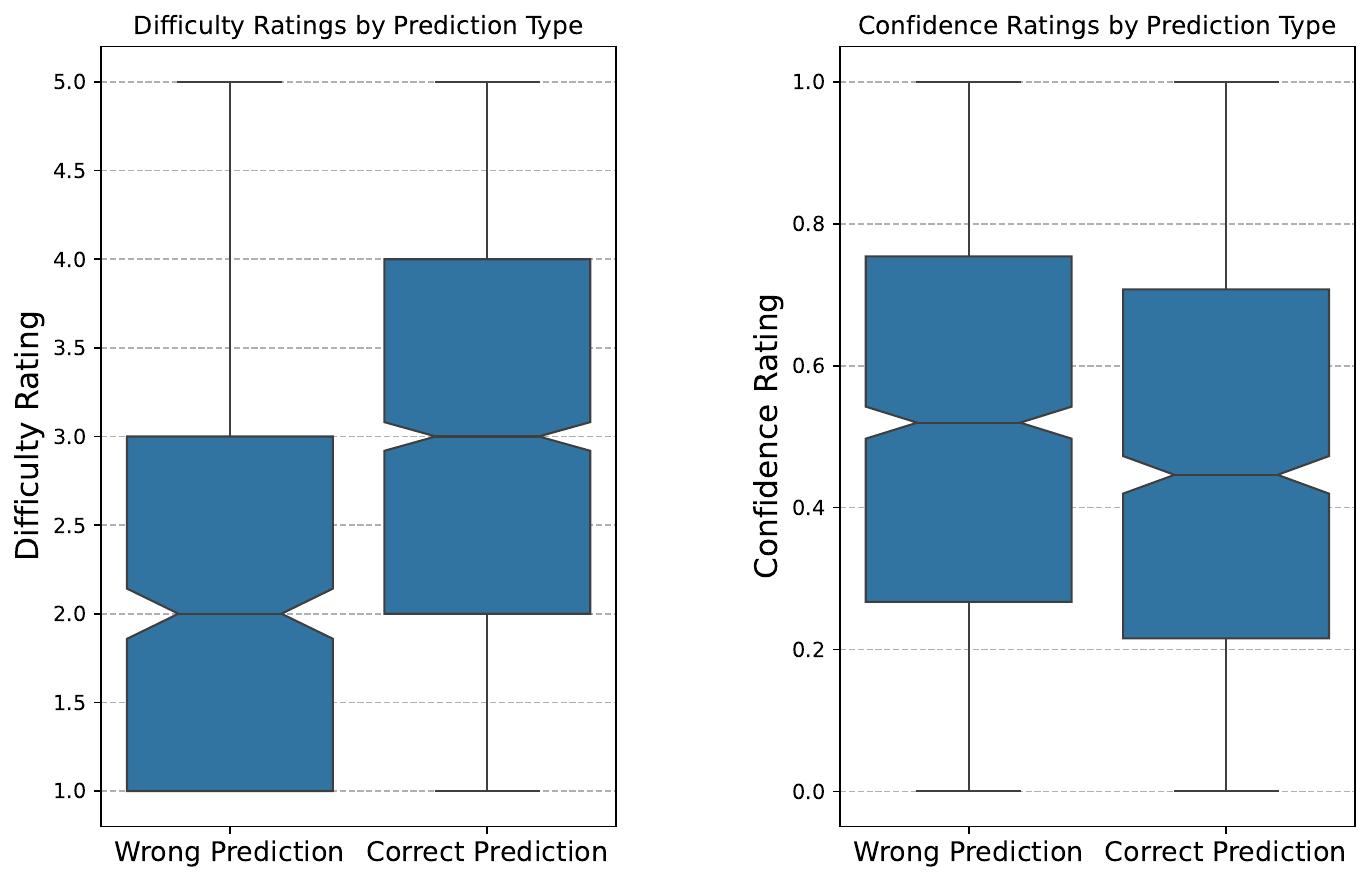}
    
    \hfill
    \begin{subfigure}[b]{0.45\linewidth}
        \centering
        
        \caption{
        }
        \label{fig:Difficulty_HoloAssist}
    \end{subfigure}
    \hfill
    \begin{subfigure}[b]{0.45\linewidth}
        \centering
        \caption{
        }
        \label{fig:Confidence_epictent}
    \end{subfigure}
     \hfill
    \caption{Distributions of difficulty ratings (a) and execution confidence ratings (b) with respect to wrong and correct predictions.}
    \label{fig:boxplots}
\end{figure}

\noindent
\textbf{Action Complexity}
We investigated how action complexity correlates with gaze-predicted mistakes in procedural tasks. We asked 40 volunteers to rate the complexity of performing actions contained in HoloAssist without looking (1 = easy, 5 = difficult). 
We then compared the complexity of the action associated to a given video segment to the ability of our model to make a correct prediction (which we term ``success'').
Results (see Figure \ref{fig:Difficulty_HoloAssist}) showed a positive correlation between difficulty and prediction success, measured with a Point Biserial Correlation of $0.3843$, with \(p < 0.05\)\footnote{We use Point Biserial Correlation as action difficulty is a continuous variable while the success of our method is a binary one.}.
This suggests that our method is particularly effective in the case of complex actions which cannot be carried out without looking, while less effective in the case of trivial tasks.

\noindent
\textbf{Confidence Level}
On the EPIC-Tent dataset, we compared if the self-rated confidence score reported by camera wearers was correlated to the success of our method.
Results (see Figure~\ref{fig:Confidence_epictent}) obtained a Point Biserial Correlation of -0.1137, \(p < 0.05\) indicating a small but significant negative correlation: our method is most effective when the self-rated confidence is higher. This suggests that gaze-based analysis is more effective in the case of novices, which reported lower confidence and probably rely more on visual observations when executing their tasks.

\begin{figure}
    \centering
    
        \centering
        \includegraphics[width=0.90\linewidth]{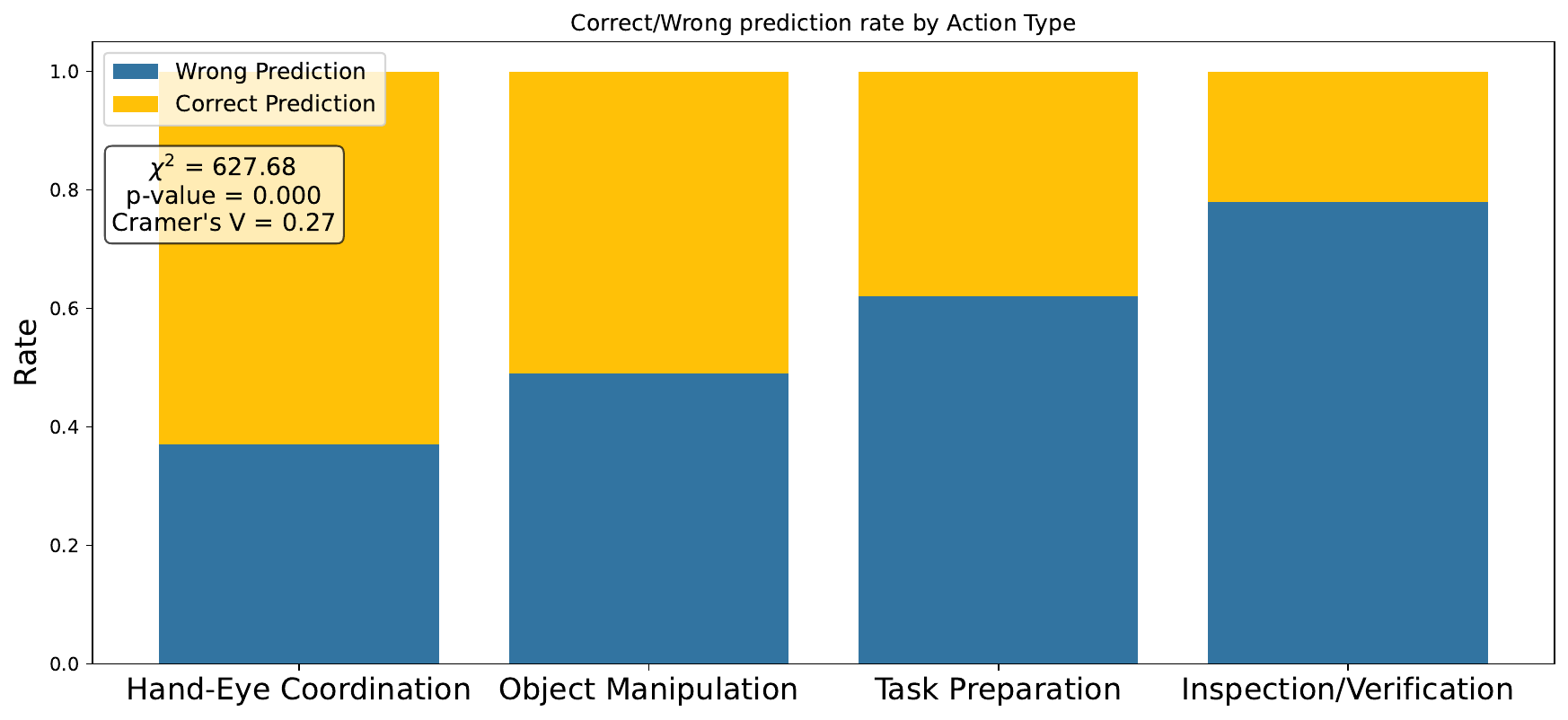}
        \caption{Distributions of Correct/Wrong pred. by action type.}
        \label{fig:prediction_by_action_type}
    
    \label{fig:uncertainty_analysis}
\end{figure}

\noindent
\textbf{Action Type} We finally assess whether the type of the performed action affects the performance of our method. To this aim, we grouped actions contained in all three datasets in four categories (Hand-Eye Coordination, Object Manipulation, Task Preparation, Inspection/Verification)\footnote{See supplementary material for more details.}.
We hence computed the number of co-ocurrences between success or failure of our method and the different action classes. Results (see Figure~\ref{fig:prediction_by_action_type}) show a Cramer's V statistic of $0.27$ (a moderate correlation of $0.27$ in a $0-1$ scale) with a p-value $p<0.05$. Gaze-based analysis proves particularly useful in the case of actions requiring hand-eye coordination and object manipulation abilities, while less effective for generic actions such as task preparation and inspection.\footnote{See supp. for more details and per class F1 and AUC scores. Qualitative results are reported in the supplementary material}


\section{Conclusion}
We proposed to perform mistake detection in egocentric videos in an unsupervised way, leveraging gaze signals. We introduced a novel \textit{gaze completion task}, where gaze trajectories are predicted based on observed video and partial gaze data, and an approach to tackle this task. Mistake detection is performed comparing predicted trajectories with ground truth, identifying instances where gaze becomes unpredictable as potential mistakes. 
Experimental validation on EPIC-Tent, HoloAssist, and IndustReal demonstrates the efficacy of our method, surpassing traditional one-class techniques and other unsupervised mistake detection methods. 
Our method is ranked first on the HoloAssist Mistake Detection challenge. Code will be publicly shared.



\section*{Acknowledgments}
This research has been supported by the project Future Artificial Intelligence Research (FAIR) – PNRR MUR Cod. PE0000013 - CUP: E63C22001940006.

{
    \small
    \bibliographystyle{ieeenat_fullname}
    \bibliography{main}

\begin{thebibliography}{46}
\providecommand{\natexlab}[1]{#1}
\providecommand{\url}[1]{\texttt{#1}}
\expandafter\ifx\csname urlstyle\endcsname\relax
  \providecommand{\doi}[1]{doi: #1}\else
  \providecommand{\doi}{doi: \begingroup \urlstyle{rm}\Url}\fi

\bibitem[Al-Naser et~al.(2019)Al-Naser, Siddiqui, Ohashi, Ahmed, Katsuyki, Takuto, and Dengel]{Naser}
Mohammad Al-Naser, Shoaib~Ahmed Siddiqui, Hiroki Ohashi, Sheraz Ahmed, Nakamura Katsuyki, Sato Takuto, and Andreas Dengel.
\newblock Ogaze: Gaze prediction in egocentric videos for attentional object selection.
\newblock In \emph{2019 Digital Image Computing: Techniques and Applications (DICTA)}, pages 1--8. IEEE, 2019.

\bibitem[Bertasius et~al.(2021)Bertasius, Wang, and Torresani]{timesformer_bertasius2021spacetime}
Gedas Bertasius, Heng Wang, and Lorenzo Torresani.
\newblock Is space-time attention all you need for video understanding?, 2021.

\bibitem[Damen et~al.(2014)Damen, Leelasawassuk, Haines, Calway, and Mayol-Cuevas]{Damen2014YouDoID}
Dima Damen, Teesid Leelasawassuk, Osian Haines, Andrew Calway, and W. Mayol-Cuevas.
\newblock You-do, i-learn: Discovering task relevant objects and their modes of interaction from multi-user egocentric video.
\newblock In \emph{British Machine Vision Conference}, 2014.

\bibitem[Ding et~al.(2023)Ding, Sener, Ma, and Yao]{ding2023mistakeYao}
Guodong Ding, Fadime Sener, Shugao Ma, and Angela Yao.
\newblock Every mistake counts in assembly.
\newblock \emph{ArXiv}, abs/2307.16453, 2023.

\bibitem[Fathi et~al.(2012)Fathi, Li, and Rehg]{fathi2012learning}
Alireza Fathi, Yin Li, and James~M Rehg.
\newblock Learning to recognize daily actions using gaze.
\newblock In \emph{Computer Vision--ECCV 2012: 12th European Conference on Computer Vision, Florence, Italy, October 7-13, 2012, Proceedings, Part I 12}, pages 314--327. Springer, 2012.

\bibitem[Feng et~al.(2021)Feng, Hong, and Zheng]{Feng2021MISTMI}
Jianfeng Feng, Fa-Ting Hong, and Weishi Zheng.
\newblock Mist: Multiple instance self-training framework for video anomaly detection.
\newblock \emph{2021 IEEE/CVF Conference on Computer Vision and Pattern Recognition (CVPR)}, pages 14004--14013, 2021.

\bibitem[Feng et~al.(2024)Feng, Wray, Sullivan, Jang, Ludwig, Gilchrist, and Mayol-Cuevas]{feng2024struggling}
Shijia Feng, Michael Wray, Brian Sullivan, Youngkyoon Jang, Casimir J.~H. Ludwig, Iain Gilchrist, and Walterio~W. Mayol-Cuevas.
\newblock Are you struggling? dataset and baselines for struggle determination in assembly videos.
\newblock \emph{ArXiv}, abs/2402.11057, 2024.

\bibitem[Flaborea et~al.(2023)Flaborea, Collorone, di~Melendugno, D'Arrigo, Prenkaj, and Galasso]{MoCoDAD_Flaborea_2023_ICCV}
Alessandro Flaborea, Luca Collorone, Guido Maria~D'Amely di Melendugno, Stefano D'Arrigo, Bardh Prenkaj, and Fabio Galasso.
\newblock Multimodal motion conditioned diffusion model for skeleton-based video anomaly detection.
\newblock In \emph{Proceedings of the IEEE/CVF International Conference on Computer Vision (ICCV)}, pages 10318--10329, 2023.

\bibitem[Flaborea et~al.(2024)Flaborea, di~Melendugno, Plini, Scofano, De~Matteis, Furnari, Farinella, and Galasso]{flaborea2024prego}
Alessandro Flaborea, Guido Maria~D'Amely di Melendugno, Leonardo Plini, Luca Scofano, Edoardo De~Matteis, Antonino Furnari, Giovanni~Maria Farinella, and Fabio Galasso.
\newblock Prego: online mistake detection in procedural egocentric videos.
\newblock In \emph{Proceedings of the IEEE/CVF Conference on Computer Vision and Pattern Recognition}, pages 18483--18492, 2024.

\bibitem[Gao et~al.(2014)Gao, Vedula, Reiley, Ahmidi, Varadarajan, Lin, Tao, Zappella, B{\'e}jar, Yuh, et~al.]{gao2014jhu}
Yixin Gao, S~Swaroop Vedula, Carol~E Reiley, Narges Ahmidi, Balakrishnan Varadarajan, Henry~C Lin, Lingling Tao, Luca Zappella, Benjam{\i}n B{\'e}jar, David~D Yuh, et~al.
\newblock Jhu-isi gesture and skill assessment working set (jigsaws): A surgical activity dataset for human motion modeling.
\newblock In \emph{MICCAI workshop: M2cai}, 2014.

\bibitem[Ghoddoosian et~al.(2023)Ghoddoosian, Dwivedi, Agarwal, and Dariush]{Ghoddoosian_2023_ICCV}
Reza Ghoddoosian, Isht Dwivedi, Nakul Agarwal, and Behzad Dariush.
\newblock Weakly-supervised action segmentation and unseen error detection in anomalous instructional videos.
\newblock In \emph{Proceedings of the IEEE/CVF International Conference on Computer Vision (ICCV)}, pages 10128--10138, 2023.

\bibitem[Grauman et~al.(2022)Grauman, Westbury, Byrne, Chavis, Furnari, Girdhar, Hamburger, Jiang, Liu, Liu, Martin, Nagarajan, Radosavovic, Ramakrishnan, Ryan, Sharma, Wray, Xu, Xu, Zhao, Bansal, Batra, Cartillier, Crane, Do, Doulaty, Erapalli, Feichtenhofer, Fragomeni, Fu, Fuegen, Gebreselasie, Gonzalez, Hillis, Huang, Huang, Jia, Khoo, Kolar, Kottur, Kumar, Landini, Li, Li, Li, Mangalam, Modhugu, Munro, Murrell, Nishiyasu, Price, Puentes, Ramazanova, Sari, Somasundaram, Southerland, Sugano, Tao, Vo, Wang, Wu, Yagi, Zhu, Arbelaez, Crandall, Damen, Farinella, Ghanem, Ithapu, Jawahar, Joo, Kitani, Li, Newcombe, Oliva, Park, Rehg, Sato, Shi, Shou, Torralba, Torresani, Yan, and Malik]{Ego4D2022CVPR}
Kristen Grauman, Andrew Westbury, Eugene Byrne, Zachary Chavis, Antonino Furnari, Rohit Girdhar, Jackson Hamburger, Hao Jiang, Miao Liu, Xingyu Liu, Miguel Martin, Tushar Nagarajan, Ilija Radosavovic, Santhosh~Kumar Ramakrishnan, Fiona Ryan, Jayant Sharma, Michael Wray, Mengmeng Xu, Eric~Zhongcong Xu, Chen Zhao, Siddhant Bansal, Dhruv Batra, Vincent Cartillier, Sean Crane, Tien Do, Morrie Doulaty, Akshay Erapalli, Christoph Feichtenhofer, Adriano Fragomeni, Qichen Fu, Christian Fuegen, Abrham Gebreselasie, Cristina Gonzalez, James Hillis, Xuhua Huang, Yifei Huang, Wenqi Jia, Weslie Khoo, Jachym Kolar, Satwik Kottur, Anurag Kumar, Federico Landini, Chao Li, Yanghao Li, Zhenqiang Li, Karttikeya Mangalam, Raghava Modhugu, Jonathan Munro, Tullie Murrell, Takumi Nishiyasu, Will Price, Paola~Ruiz Puentes, Merey Ramazanova, Leda Sari, Kiran Somasundaram, Audrey Southerland, Yusuke Sugano, Ruijie Tao, Minh Vo, Yuchen Wang, Xindi Wu, Takuma Yagi, Yunyi Zhu, Pablo Arbelaez, David Crandall, Dima Damen, Giovanni~Maria
  Farinella, Bernard Ghanem, Vamsi~Krishna Ithapu, C.~V. Jawahar, Hanbyul Joo, Kris Kitani, Haizhou Li, Richard Newcombe, Aude Oliva, Hyun~Soo Park, James~M. Rehg, Yoichi Sato, Jianbo Shi, Mike~Zheng Shou, Antonio Torralba, Lorenzo Torresani, Mingfei Yan, and Jitendra Malik.
\newblock Ego4d: Around the {W}orld in 3,000 {H}ours of {E}gocentric {V}ideo.
\newblock In \emph{IEEE/CVF Computer Vision and Pattern Recognition (CVPR)}, 2022.

\bibitem[Huang et~al.(2018)Huang, Cai, Li, and Sato]{Huang2018PredictingGI}
Yifei Huang, Minjie Cai, Zhenqiang Li, and Yoichi Sato.
\newblock Predicting gaze in egocentric video by learning task-dependent attention transition.
\newblock \emph{ArXiv}, abs/1803.09125, 2018.

\bibitem[Jang et~al.(2019{\natexlab{a}})Jang, Sullivan, Ludwig, Gilchrist, Damen, and Mayol-Cuevas]{EPIC-Tent9022634}
Youngkyoon Jang, Brian Sullivan, Casimir Ludwig, Iain~D. Gilchrist, Dima Damen, and Walterio Mayol-Cuevas.
\newblock Epic-tent: An egocentric video dataset for camping tent assembly.
\newblock In \emph{2019 IEEE/CVF International Conference on Computer Vision Workshop (ICCVW)}, pages 4461--4469, 2019{\natexlab{a}}.

\bibitem[Jang et~al.(2019{\natexlab{b}})Jang, Sullivan, Ludwig, Gilchrist, Damen, and Mayol-Cuevas]{Tent}
Youngkyoon Jang, Brian Sullivan, Casimir Ludwig, Iain~D. Gilchrist, Dima Damen, and Walterio Mayol-Cuevas.
\newblock Epic-tent: An egocentric video dataset for camping tent assembly.
\newblock In \emph{2019 IEEE/CVF International Conference on Computer Vision Workshop (ICCVW)}, pages 4461--4469, 2019{\natexlab{b}}.

\bibitem[Konrad et~al.(2024)Konrad, Padmanaban, Buckmaster, Boyle, and Wetzstein]{konrad2024gazegpt}
Robert Konrad, Nitish Padmanaban, J.~Gabriel Buckmaster, Kevin~C. Boyle, and Gordon Wetzstein.
\newblock Gazegpt: Augmenting human capabilities using gaze-contingent contextual ai for smart eyewear, 2024.

\bibitem[Koochaki and Najafizadeh(2018)]{Koochaki}
Fatemeh Koochaki and Laleh Najafizadeh.
\newblock Predicting intention through eye gaze patterns.
\newblock \emph{2018 IEEE Biomedical Circuits and Systems Conference (BioCAS)}, pages 1--4, 2018.

\bibitem[Krafka et~al.(2016)Krafka, Khosla, Kellnhofer, Kannan, Bhandarkar, Matusik, and Torralba]{Krafka}
Kyle Krafka, Aditya Khosla, Petr Kellnhofer, Harini Kannan, Suchendra~M. Bhandarkar, Wojciech Matusik, and Antonio Torralba.
\newblock Eye tracking for everyone.
\newblock \emph{2016 IEEE Conference on Computer Vision and Pattern Recognition (CVPR)}, pages 2176--2184, 2016.

\bibitem[Lai et~al.(2022)Lai, Liu, Ryan, and Rehg]{lai2022eye}
Bolin Lai, Miao Liu, Fiona Ryan, and James Rehg.
\newblock In the eye of transformer: Global-local correlation for egocentric gaze estimation.
\newblock \emph{British Machine Vision Conference}, 2022.

\bibitem[Land et~al.(1999)Land, Mennie, and Rusted]{gaze1999}
M. Land, Neil Mennie, and Jennifer Rusted.
\newblock The roles of vision and eye movements in the control of activities of daily living.
\newblock \emph{Perception}, 28:\penalty0 1311--28, 1999.

\bibitem[Li et~al.(2013)Li, Fathi, and Rehg]{li2013learning}
Yin Li, Alireza Fathi, and James~M Rehg.
\newblock Learning to predict gaze in egocentric video.
\newblock In \emph{Proceedings of the IEEE international conference on computer vision}, pages 3216--3223, 2013.

\bibitem[Li et~al.(2018)Li, Liu, and Rehg]{Li_2018_ECCV}
Yin Li, Miao Liu, and James~M. Rehg.
\newblock In the eye of beholder: Joint learning of gaze and actions in first person video.
\newblock In \emph{Proceedings of the European Conference on Computer Vision (ECCV)}, 2018.

\bibitem[Liu et~al.(2021)Liu, Nie, Long, Zhang, and Li]{liu2021hf2vad}
Zhian Liu, Yongwei Nie, Chengjiang Long, Qing Zhang, and Guiqing Li.
\newblock A hybrid video anomaly detection framework via memory-augmented flow reconstruction and flow-guided frame prediction.
\newblock In \emph{Proceedings of the IEEE International Conference on Computer Vision}, 2021.

\bibitem[Lu et~al.(2013)Lu, Shi, and Jia]{6751449}
Cewu Lu, Jianping Shi, and Jiaya Jia.
\newblock Abnormal event detection at 150 fps in matlab.
\newblock In \emph{2013 IEEE International Conference on Computer Vision}, pages 2720--2727, 2013.

\bibitem[Lv et~al.(2023)Lv, Yue, Sun, Luo, Cui, and Zhang]{Lv2023UnbiasedMI}
Hui Lv, Zhongqi Yue, Qianru Sun, Bin Luo, Zhen Cui, and Hanwang Zhang.
\newblock Unbiased multiple instance learning for weakly supervised video anomaly detection.
\newblock \emph{2023 IEEE/CVF Conference on Computer Vision and Pattern Recognition (CVPR)}, pages 8022--8031, 2023.

\bibitem[Masuda et~al.(2020)Masuda, Hachiuma, Fujii, and Saito]{Masuda}
Mana Masuda, Ryo Hachiuma, Ryo Fujii, and Hideo Saito.
\newblock \emph{Unsupervised Anomaly Detection of the First Person in Gait from an Egocentric Camera}, page 604–617.
\newblock Springer-Verlag, Berlin, Heidelberg, 2020.

\bibitem[Mazzamuto et~al.(2024)Mazzamuto, Ragusa, Furnari, and Farinella]{mazzamuto_gaze}
Michele Mazzamuto, Francesco Ragusa, Antonino Furnari, and Giovanni~Maria Farinella.
\newblock Learning to detect attended objects in cultural sites with gaze signals and weak object supervision.
\newblock \emph{J. Comput. Cult. Herit.}, 2024.

\bibitem[Oquab et~al.(2023)Oquab, Darcet, Moutakanni, Vo, Szafraniec, Khalidov, Fernandez, Haziza, Massa, El-Nouby, et~al.]{oquab2023dinov2}
Maxime Oquab, Timoth{\'e}e Darcet, Th{\'e}o Moutakanni, Huy Vo, Marc Szafraniec, Vasil Khalidov, Pierre Fernandez, Daniel Haziza, Francisco Massa, Alaaeldin El-Nouby, et~al.
\newblock Dinov2: Learning robust visual features without supervision.
\newblock \emph{arXiv preprint arXiv:2304.07193}, 2023.

\bibitem[Peacock et~al.(2022)Peacock, Lafreniere, Zhang, Santosa, Benko, and Jonker]{Peacock}
Candace~E. Peacock, Ben Lafreniere, Ting Zhang, Stephanie Santosa, Hrvoje Benko, and Tanya~R. Jonker.
\newblock Gaze as an indicator of input recognition errors.
\newblock \emph{Proc. ACM Hum.-Comput. Interact.}, 6\penalty0 (ETRA), 2022.

\bibitem[Pelz and Canosa(2001)]{Pelz}
Jeff Pelz and Roxanne Canosa.
\newblock Oculomotor behavior and perceptual strategies in complex tasks.
\newblock \emph{Vision research}, 41:\penalty0 3587--96, 2001.

\bibitem[Sakoe(1978)]{Sakoe1978DynamicPA}
Hiroaki Sakoe.
\newblock Dynamic programming algorithm optimization for spoken word recognition.
\newblock \emph{IEEE Transactions on Acoustics, Speech, and Signal Processing}, 26:\penalty0 159--165, 1978.

\bibitem[Schoonbeek et~al.(2024)Schoonbeek, Houben, Onvlee, van~der Sommen, et~al.]{schoonbeek2024industreal}
Tim~J Schoonbeek, Tim Houben, Hans Onvlee, Fons van~der Sommen, et~al.
\newblock Industreal: A dataset for procedure step recognition handling execution errors in egocentric videos in an industrial-like setting.
\newblock In \emph{Proceedings of the IEEE/CVF Winter Conference on Applications of Computer Vision}, pages 4365--4374, 2024.

\bibitem[Seminara et~al.(2024)Seminara, Farinella, and Furnari]{seminara2024differentiable}
Luigi Seminara, Giovanni~Maria Farinella, and Antonino Furnari.
\newblock Differentiable task graph learning: Procedural activity representation and online mistake detection from egocentric videos.
\newblock In \emph{Advances in Neural Information Processing Systems}, 2024.

\bibitem[Sener et~al.(2022)Sener, Chatterjee, Shelepov, He, Singhania, Wang, and Yao]{sener2022assembly101}
Fadime Sener, Dibyadip Chatterjee, Daniel Shelepov, Kun He, Dipika Singhania, Robert Wang, and Angela Yao.
\newblock Assembly101: A large-scale multi-view video dataset for understanding procedural activities.
\newblock In \emph{Proceedings of the IEEE/CVF Conference on Computer Vision and Pattern Recognition}, pages 21096--21106, 2022.

\bibitem[Singhania et~al.(2023)Singhania, Rahaman, and Yao]{singhania2023c2f}
Dipika Singhania, Rahul Rahaman, and Angela Yao.
\newblock C2f-tcn: A framework for semi-and fully-supervised temporal action segmentation.
\newblock \emph{IEEE Transactions on Pattern Analysis and Machine Intelligence}, 45\penalty0 (10):\penalty0 11484--11501, 2023.

\bibitem[Sood et~al.(2020)Sood, Tannert, Mueller, and Bulling]{NEURIPS2020_460191c7}
Ekta Sood, Simon Tannert, Philipp Mueller, and Andreas Bulling.
\newblock Improving natural language processing tasks with human gaze-guided neural attention.
\newblock In \emph{Advances in Neural Information Processing Systems}, 2020.

\bibitem[Soran et~al.(2015)Soran, Farhadi, and Shapiro]{Lights_Off}
Bilge Soran, Ali Farhadi, and Linda Shapiro.
\newblock Generating notifications for missing actions: Don't forget to turn the lights off!
\newblock In \emph{2015 IEEE International Conference on Computer Vision (ICCV)}, pages 4669--4677, 2015.

\bibitem[Steil et~al.(2019)Steil, Koelle, Heuten, Boll, and Bulling]{steil2019privaceye}
Julian Steil, Marion Koelle, Wilko Heuten, Susanne Boll, and Andreas Bulling.
\newblock Privaceye: privacy-preserving head-mounted eye tracking using egocentric scene image and eye movement features.
\newblock In \emph{Proceedings of the 11th ACM symposium on eye tracking research \& applications}, pages 1--10, 2019.

\bibitem[Stergiou et~al.(2024)Stergiou, De~Weerdt, and Deligiannis]{TrajREC_stergiou2024holistic}
Alexandros Stergiou, Brent De~Weerdt, and Nikos Deligiannis.
\newblock Holistic representation learning for multitask trajectory anomaly detection.
\newblock In \emph{IEEE/CVF Winter Conference on Applications of Computer Vision (WACV)}, 2024.

\bibitem[Sultani et~al.(2018)Sultani, Chen, and Shah]{Sultani_2018_CVPR}
Waqas Sultani, Chen Chen, and Mubarak Shah.
\newblock Real-world anomaly detection in surveillance videos.
\newblock In \emph{The IEEE Conference on Computer Vision and Pattern Recognition (CVPR)}, 2018.

\bibitem[Tavakoli et~al.(2019)Tavakoli, Rahtu, Kannala, and Borji]{8658619}
Hamed~Rezazadegan Tavakoli, Esa Rahtu, Juho Kannala, and Ali Borji.
\newblock Digging deeper into egocentric gaze prediction.
\newblock In \emph{2019 IEEE Winter Conference on Applications of Computer Vision (WACV)}, pages 273--282, 2019.

\bibitem[Wang et~al.(2023)Wang, Kwon, Rad, Pan, Chakraborty, Andrist, Bohus, Feniello, Tekin, Frujeri, Joshi, and Pollefeys]{HoloAssist2023}
Xin Wang, Taein Kwon, Mahdi Rad, Bowen Pan, Ishani Chakraborty, Sean Andrist, Dan Bohus, Ashley Feniello, Bugra Tekin, Felipe~Vieira Frujeri, Neel Joshi, and Marc Pollefeys.
\newblock Holoassist: an egocentric human interaction dataset for interactive ai assistants in the real world.
\newblock In \emph{Proceedings of the IEEE/CVF International Conference on Computer Vision (ICCV)}, pages 20270--20281, 2023.

\bibitem[Xu et~al.(2015)Xu, Ricci, Yan, Song, and Sebe]{Xu2015LearningDR}
Dan Xu, Elisa Ricci, Yan Yan, Jingkuan Song, and N. Sebe.
\newblock Learning deep representations of appearance and motion for anomalous event detection.
\newblock In \emph{British Machine Vision Conference}, 2015.

\bibitem[Yao et~al.(2023)Yao, Ye, He, and Elsayed]{10204584}
Yushi Yao, Chang Ye, Junfeng He, and Gamaleldin~F. Elsayed.
\newblock Teacher-generated spatial-attention labels boost robustness and accuracy of contrastive models.
\newblock In \emph{2023 IEEE/CVF Conference on Computer Vision and Pattern Recognition (CVPR)}, pages 23282--23291, 2023.

\bibitem[Zheng et~al.(2022)Zheng, Yang, Mo, Li, Yu, Liu, Liu, and Guibas]{Zheng2022GIMOGH}
Yang Zheng, Yanchao Yang, Kaichun Mo, Jiaman Li, Tao Yu, Yebin Liu, Karen Liu, and Leonidas~J. Guibas.
\newblock Gimo: Gaze-informed human motion prediction in context.
\newblock In \emph{European Conference on Computer Vision}, 2022.

\bibitem[Zhou et~al.(2024)Zhou, Liu, and Gou]{zhou2024learningobservergazezeroshotattention}
Yuchen Zhou, Linkai Liu, and Chao Gou.
\newblock Learning from observer gaze:zero-shot attention prediction oriented by human-object interaction recognition, 2024.

\end{thebibliography}
}

\setcounter{page}{1}
\maketitlesupplementary

%
%
%

\appendix

\section{Implementation details}

Our model implementation is primarily based on the approach outlined in~\cite{lai2022eye}, with hyperparameters adjusted accordingly. Below, we elaborate on the key aspects of our implementation, highlighting the differences and specific choices made to enhance the model's performance.
\\

\noindent
\textbf{Code Availability:} To ensure reproducibility and provide further implementation details, we will share the complete codebase upon publication.

\begin{figure}[tb]
  \centering
  \includegraphics[width=\linewidth]{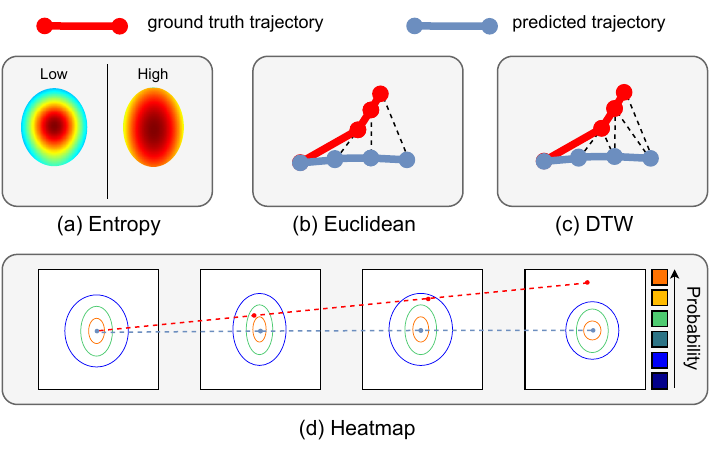}
  \caption{We consider three approaches to compare the ground truth with respect to the predicted trajectories in order to determine a mistake. (a) Entropy.  (b) Euclidean distance between the two trajectories. (c) Dynamic Time Warping (DTW). (d) Average value of the ground truth trajectory at the predicted heatmaps.}
  \label{fig:scoring_functions}
\end{figure}

\noindent
\textbf{Stride Adjustment:} In contrast to the stride of 8 used in~\cite{lai2022eye}, we opted for a stride of 1 during training. This modification allows the model to process consecutive frames without skipping, leading to finer granularity in temporal feature extraction. Our experiments indicated that this adjustment results in marginal improvements in both gaze estimation and mistake prediction accuracy.\\
\noindent
\textbf{Overfitting Prevention:} To mitigate the risk of overfitting, we incorporated a weight decay parameter set to 0.07. This regularization technique helps in controlling the complexity of the model by penalizing large weights, thereby promoting generalization to unseen data.\\
\noindent
\textbf{Batch Size and Frame Processing:} We configured the batch size to process 4 clips, each containing 8 frames. Specifically, our approach involves processing each video using non-overlapping windows of 8 consecutive frames. Consequently, each batch comprises 4 such windows, totaling 32 frames per batch (i.e., \(8 \text{ frames/clip} \times 4 \text{ clips/batch} = 32 \text{ frames/batch}\)). This setting ensures that the model captures sufficient temporal context while maintaining manageable memory usage.\\
\noindent
\textbf{Training Loss}
 Following~\cite{lai2022eye}, we consider gaze prediction as defining a probability distribution over the 2D image plane of each input frame. 
 Our proposed method leverages an architecture modified for gaze completion to predict missing segments of gaze trajectories.
Ablations on single frames showed that looking at sequences of frames, which create a trajectory, is more effective for detecting anomalies due to the importance of changes over time.
We train the model by minimizing the sum of the Kullback–Leibler divergence between the predicted gaze maps $\hat P(i)$ and the ground truth ones $Q(i)$ at each frame $i$:

\begin{equation}
 L_{\text{KL}}(\hat P \parallel Q) = \sum_{i} \hat P(i) \log \left( \frac{\hat P(i)}{Q(i)} \right)
\end{equation}
\noindent
\textbf{Graphical Illustration of scoring functions}
Figure~\ref{fig:scoring_functions} illustrates the scoring function considered in this study. Entropy is the only scoring function which does not require any ground truth gaze as input, but only evaluates the level of uncertainty of the predicted heatmaps. The Euclidean and DTW scoring functions compute two forms of distances between predicted and ground truth trajectories. The heatmap scoring function evaluates the probability of predicted gaze under the points indicated by the ground truth trajectory. The heatmap scoring function achieves best results in our experiments. The main paper reports the formal definition of such scoring functions.
\paragraph{MoCoDAD Baseline}
Following methodologies from \cite{MoCoDAD_Flaborea_2023_ICCV}, we employ a sliding window approach to segment each agent’s gaze/hands history. A window size of 8 frames is utilized, with the initial 4 frames dedicated to condition setting and the subsequent frames for the diffusion process. Hyperparameters are set as $\lambda_1 = \lambda_2 = 1$. Training proceeds end-to-end using the Adam optimizer with a learning rate of $1 \times 10^{-4}$, employing exponential decay over 25 epochs. The diffusion process utilizes $\beta_1 = 1 \times 10^{-4}$, $\beta_T = 2 \times 10^{-2}$ for $T = 10$, and incorporates the cosine variance scheduler.

\paragraph{TrajREC Baseline}

We followed the implementation proposed in \textit{TrajREC}~\cite{TrajREC_stergiou2024holistic} official code release\footnote{\url{https://github.com/alexandrosstergiou/TrajREC}}, adapting it for gaze/hands trajectory analysis. The approach encodes temporally occluded gaze/hands trajectories, jointly learns latent representations of occluded segments, and reconstructs trajectories based on expected motions across different temporal segments.

For both methods, if a frame does not contain gaze or hand keypoints, we exclude that frame from the score calculation for the segment.

\begin{table}[t]

    \centering
        \centering

        \begin{threeparttable}
        \resizebox{0.99\linewidth}{!}{
            \begin{tabular}{lcccc}
                \hline
                \textbf{Method} & \textbf{Fusion}& \textbf{F1} & \textbf{Recall} & \textbf{Precision} \\
                \hline
                Gaze Prediction & //& 0.37 & 0.65 & 0.26 \\
                Gaze Completion & CH & 0.38 & 0.67 & 0.29 \\
                Gaze Completion & CH + CORR & \textbf{0.40} & \textbf{0.70} & \textbf{0.31} \\
                \hline
            \end{tabular}
        }
        \end{threeparttable}
      \caption{Comparison of GLC and the proposed Gaze Completion approach for Gaze Estimation on EPIC-Tent.}
 \label{tab:GLC_Gaze_Estimation_Epic}
    
\end{table}

\paragraph{Action Type Classification}
To assess whether the type of performed action affects the performance of our method, we grouped actions contained in all three datasets into four categories: \textit{Hand-Eye Coordination}, \textit{Object Manipulation}, \textit{Task Preparation}, and \textit{Inspection/Verification}. For categorization, we prompted GPT-4 with a full list of actions using the following prompt:

\textit{In the context of how gaze affects actions, organize the following actions into groups that align with gaze literature. Group the actions from those that involve the most fine-grained gaze coordination to those that involve less gaze precision.}

The list was then manually revised. The full classification is shown in Table \ref{tab:action_classification}

\renewcommand{\arraystretch}{1.5} 
\begin{table*}[h!]
    \centering
    \resizebox{\textwidth}{!}{ 
    \begin{tabular}{|c|c|c|}
\hline
\textbf{Category}                        & \textbf{Dataset} & \textbf{Actions}                                                                                 \\ \hline
\multirow{3}{*}{Hand-Eye Coordination } & EpicTent   & assemble, insert stake, insert support, insert support tab, tie top                                               \\ \cline{2-3} 
                                         & HoloAssist       & touch, place, lift, press, flip, unscrew, rotate, slide, insert, close, turn, screw, disassemble \\ \cline{2-3} 
                                         & IndustReal       & fit, plug, tighten, loosen                                                                       \\ \hline\hline
\multirow{3}{*}{Object Manipulation}     & EpicTent         & spread tent, place guyline                                                                       \\ \cline{2-3} 
                                       & HoloAssist & adjust, empty, drop, clean, make, pour, split, mix-stir, stack-pile, load, mount, lock, unlock, shift, grab, pull \\ \cline{2-3} 
                                         & IndustReal       & put, take, pull                                                                                  \\ \hline\hline
\multirow{3}{*}{Task Preparation}        & EpicTent         & pickup/open stakebag, pickup/open supportbag, pickup/open tentbag                                \\ \cline{2-3} 
                                         & HoloAssist       & withdraw, exchange, hold, break, approach, stand, align                                          \\ \cline{2-3} 
                                         & IndustReal       & align                                                                                            \\ \hline\hline
\multirow{3}{*}{Inspection/Verification} & EpicTent         & instruction, place ventcover                                                                     \\ \cline{2-3} 
                                         & HoloAssist       & inspect, validate, point, tap, click, push                                                       \\ \cline{2-3} 
                                         & IndustReal       & check, browse                                                                                    \\ \hline
\end{tabular}%
}
    \caption{Classification of actions by category across datasets based on gaze involvement.}
    \label{tab:action_classification}
\end{table*}
\renewcommand{\arraystretch}{1}

\begin{table*}[t]
    \centering
    \begin{threeparttable}
        \resizebox{0.99\linewidth}{!}{
            \begin{tabular}{lccccccccccc}
                \hline
                \textbf{Method} & \textbf{Sup. Level} & \textbf{Overall F1} & \textbf{Overall AUC} & \multicolumn{2}{|c|}{\textbf{Hand-Eye Coord.}} & \multicolumn{2}{c|}{\textbf{Object Manip.}} & \multicolumn{2}{c|}{\textbf{Task Prep.}} & \multicolumn{2}{c}{\textbf{Inspect/Verif.}} \\
                \cline{5-12}
                & & & & F1 & AUC & F1 & AUC & F1 & AUC & F1 & AUC \\
                \hline
                Random & // & 0.36 & 0.51 & -- & -- & -- & -- & -- & -- & -- & -- \\
                TimeSformer~\cite{timesformer_bertasius2021spacetime} & Fully Supervised & \underline{0.49} & \underline{0.67} & 0.452 & 0.615 & 0.474 & 0.636 & 0.551 & 0.691 & 0.532 & 0.678 \\
                C2F~\cite{singhania2023c2f} & Fully Supervised & \textbf{0.58} & \textbf{0.72} & 0.506 & 0.600  & 0.5622 & 0.771 & 0.5138 & 0.686 & 0.741 & 0.857 \\ 
                
                \hline
                GLC~\cite{lai2022eye} & One-Class & 0.46 & 0.66 & 0.524 & 0.704 & 0.495 & 0.665 & 0.425 & 0.579 & 0.396 & 0.556 \\
                Ours & One-Class & \underline{0.52} & \underline{0.69} & 0.741 & 0.839 & 0.612 & 0.734 & 0.489 & 0.643 & 0.244 & 0.543 \\
                Ours + MoCoDAD (H)* & One-Class & \textbf{0.54} & \textbf{0.72} & 0.753 & 0.872 & 0.631 & 0.764 & 0.498 & 0.657 & 0.257 & 0.543 \\
                \hline
                GLC~\cite{lai2022eye} & Unsupervised & 0.44 & 0.61 & 0.542 & 0.694 & 0.474 & 0.657 & 0.406 & 0.563 & 0.338 & 0.526 \\
                Ours & Unsupervised & \underline{0.51} & \underline{0.69} & 0.711 & 0.839 & 0.603 & 0.723 & 0.483 & 0.637 & 0.240 & 0.531 \\
                Ours + MoCoDAD (H)* & Unsupervised & \textbf{0.52} & \textbf{0.70} & 0.714 & 0.862 & 0.602 & 0.754 & 0.489 & 0.646 & 0.253 & 0.535 \\
                \hline
            \end{tabular}
            
        }
        \begin{tablenotes}
            \item[*] \footnotesize Late fusion
        \end{tablenotes}
    \end{threeparttable}
    \caption{Mistake detection results on EPIC-Tent by category. Best results are \textbf{in bold}, second best results are \underline{underlined}.}
    \label{tab:GLC_EPIC_ACTION_TYPE}
\end{table*}

\begin{table}[]
    \centering
    \resizebox{\columnwidth}{!}{
           \begin{tabular}{|c|c|c|c|c|}
            \hline
            \textbf{Baseline} &\textbf{Future frames} & \textbf{F1} &  \textbf{Precision} &  \textbf{Recall} \\ \hline
            Gaze Completion & 1&0.46 & 0.39 & 0.59 \\ \hline
            Gaze Completion & 2&0.47 & 0.38 & 0.62 \\ \hline
            Gaze Completion & 3&0.47 & 0.37 & 0.63 \\ \hline
            Gaze Completion & 4&0.49 & 0.34 & 0.88 \\ 
            \hline
        \end{tabular}}
    \caption{Performance ablation for different prediction lengths. Smaller windows yield higher precision but lower recall. The best F1 score is achieved when predicting 4 frames into the future.}
    \label{tab:prediction_lengths}
\end{table}

\section{Additional Ablations}
This section reports additional ablations which could not be included in the submitted paper due to space limits.
\\
\noindent

\subsection{Performance Comparison Across Action Types}

Table \ref{tab:GLC_EPIC_ACTION_TYPE} compares the performance of the considered baselines and our proposed method across different action types.

The fully supervised C2F method achieves overall F1 and AUC scores of 0.58 and 0.72, respectively, maintaining stable performance across the four action types. The best performance is observed in \textit{Inspect/Verify} actions (F1: 0.74, AUC: 0.85), likely due to the strong visual cues inherent to these tasks (e.g., instruction sheets).

In contrast, under both \textit{One-Class} and \textit{Unsupervised} scenarios, the proposed gaze-based approaches show varying performance depending on the action type. Stronger results are observed in tasks requiring \textit{Hand-Eye Coordination} and \textit{Object Manipulation} skills. Under the \textit{One-Class} supervision level, our method achieves an F1 score of 0.741 and an AUC of 0.839 for hand-eye coordination tasks (\textcolor{ForestGreen}{+21\%} vs Overall AUC). This indicates its effectiveness in learning ``normal" attention patterns and detecting mistakes during complex actions where gaze and motor coordination are crucial.

Conversely, for simpler actions, such as \textit{Task Preparation} and \textit{Inspect/Verify}, the proposed approaches are less effective. This is likely due to the high gaze variability inherent in less skill-intensive tasks. For instance, under the \textit{One-Class} supervision level, our method achieves an F1 score of 0.257 and an AUC of 0.543 (\textcolor{Bittersweet}{-24\%} vs Overall AUC) for task preparation actions.

\subsubsection{Performance of the proposed gaze completion approach vs standard gaze prediction}

Results in main paper Table~1 compared the performance of the proposed mistake detection method based on gaze completion versus different methods, including a baseline method based on the standard gaze prediction task implemented with the method of~\cite{lai2022eye}.
In Table~\ref{tab:GLC_Gaze_Estimation_Epic}, we instead compare the performance of the proposed gaze completion approach with standard gaze prediction based on~\cite{lai2022eye} on the EPIC-Tent dataset. Just using channel fusion brings a performance boost, achieving an F1-score of $0.38$, recall of $0.67$, and precision of $0.29$, while combining channel and correlation fusion brings best results with an F1-score of $0.40$, recall of $0.70$, and precision of $0.31$, suggesting that conditioning on partial trajectories makes gaze prediction less uncertain and the proposed approach can leverage the informative content provided by the input trajectory surpassing the performance of standard gaze prediction.
Moreover, the performance in Table~\ref{tab:GLC_Gaze_Estimation_Epic} correlates with the results in Table~1 of the main paper, suggesting that accurate gaze prediction enhances mistake detection performance. Specifically, the proposed approach excels in gaze prediction for ``Correct execution" frames, although it loses accuracy for ``Mistake" frames. Given that ``Correct execution" frames are generally more frequent, the F1 score improves overall, but the gap in prediction accuracy between ``Correct execution" and ``Mistake" frames widens. This discrepancy, however, benefits trajectory-based comparisons in mistake detection, as the increased accuracy in ``Correct execution" frames helps to better identify errors in subsequent frames.
\\

\subsection{Length of prediction and performance}
Table \ref{tab:prediction_lengths} ablates performance for different prediction lengths. Smaller windows lead to higher precision due to short future trajectories being more predictable, but also lower recall, with the best F1 score when predicting 4 frames into the future.\\
As the prediction window extends from 1 to 4 frames, the model's recall improves, indicating more mistakes detected. However, this is offset by a reduction in precision, leading to more false positives.

\begin{figure}[tb]
  \centering
  \includegraphics[width=\linewidth]{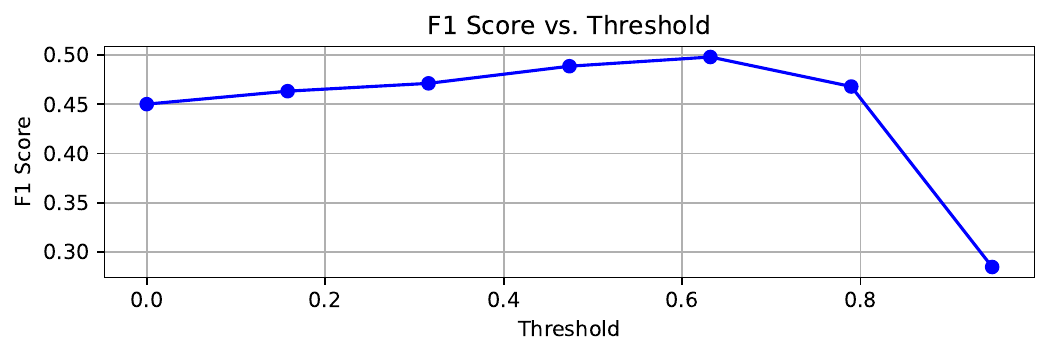}
  \caption{Length of prediction.}
  \label{fig:scoring}
\end{figure}



\subsection{Chosen thresholds and sensitivity} 
We report F1 scores obtained at each method's optimal thresholds, which we'll report in the paper. Figure~\ref{fig:scoring} shows how the F1 score of our best method (\textit{Unsupervised - Ours}, Table 2 of main paper) changes when varying the threshold. Performance is stable for a range of threshold values.

\begin{figure*}[htbp]
    \centering
    \begin{subfigure}[b]{\textwidth}
        \centering
        \includegraphics[width=0.99\textwidth]{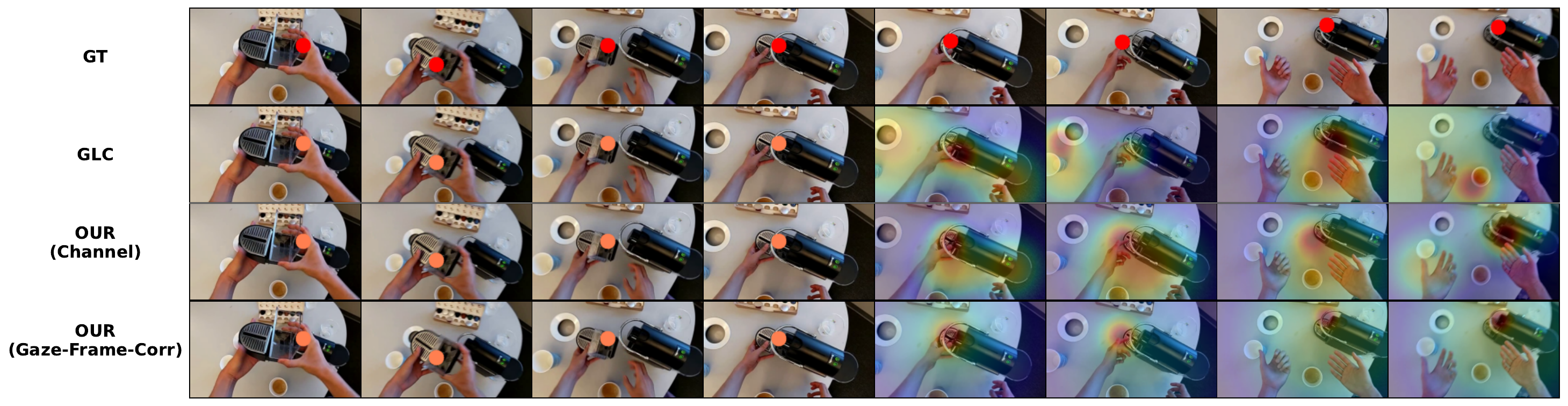}
        \includegraphics[width=0.99\textwidth]{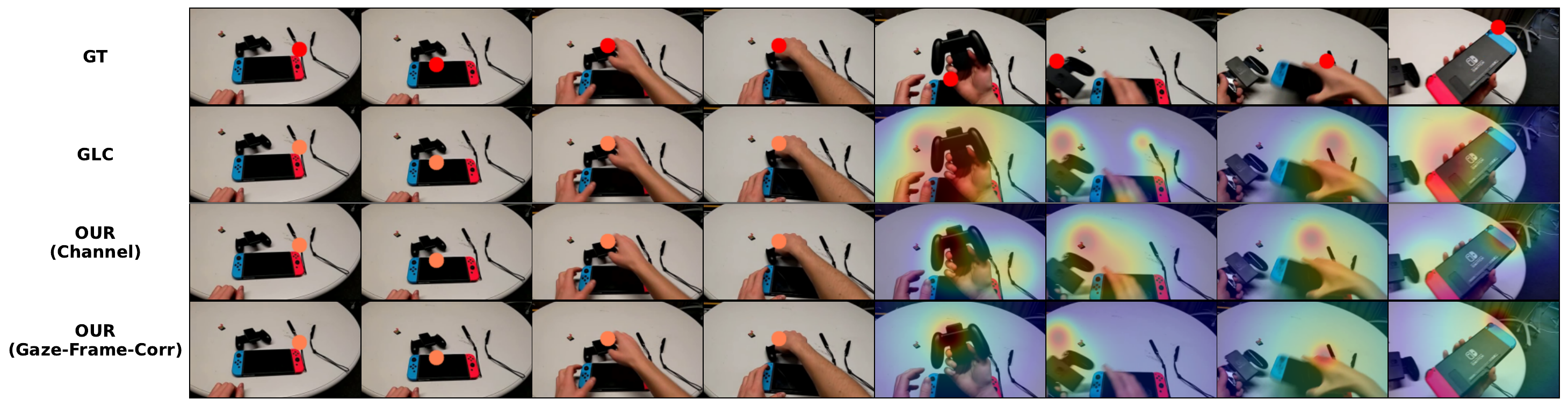}
        \caption{Correct prediction of Correct action.}
        \label{fig:qualitative_big_correct}
    \end{subfigure}
    
    \vspace{50pt} 

    \begin{subfigure}[b]{\textwidth}
        \centering
        \includegraphics[width=0.99\textwidth]{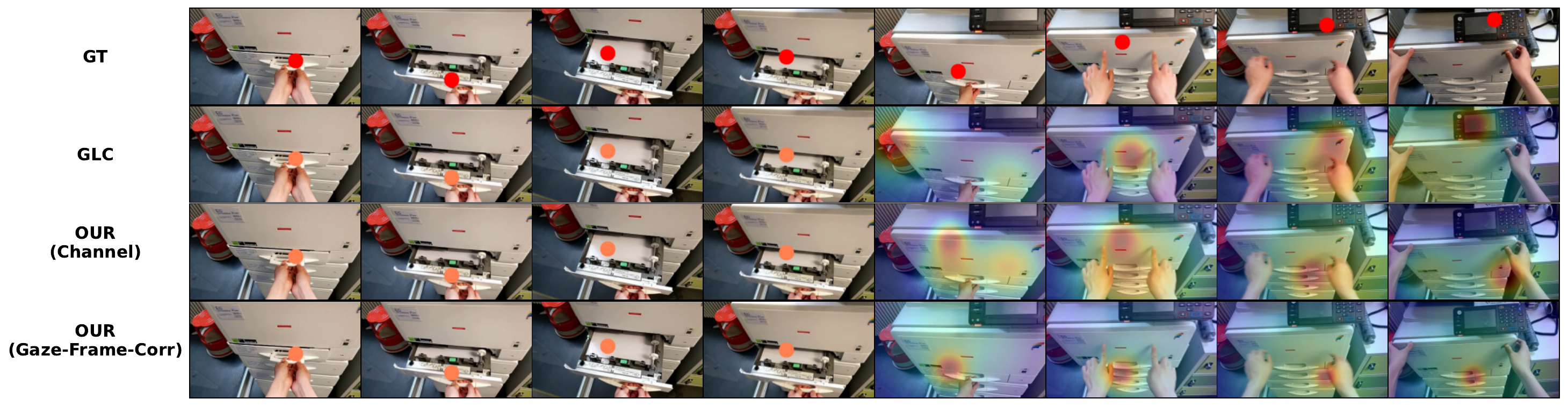}
        \includegraphics[width=0.99\textwidth]{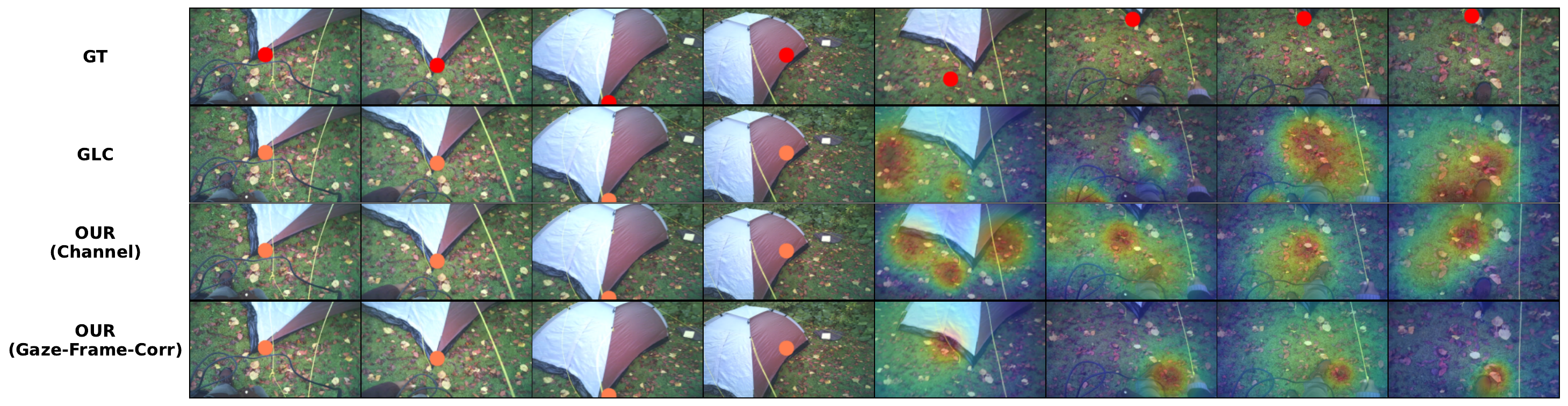}
        \caption{Correct prediction of Mistake Action.}
        \label{fig:qualitative_big_mistake}
    \end{subfigure}
    
    \caption{Qualitative examples. The first four columns represent the inputs (with the input gaze 2D points highlighted in orange). The latter four columns show the predicted outputs in the form of heatmaps.}
    \label{fig:qualitative_big}
\end{figure*}

\subsection{Qualitative Results and Failure Cases}



Figure~\ref{fig:qualitative_big} illustrates the performance of various baselines and the proposed approach for both correct predictions in \textit{Correct Execution} cases (a) and in \textit{Incorrect Execution} cases (b). The top row displays the ground truth, followed by predictions from the GLC method, our proposed ``Channel" approach, and, at the bottom, the "Gaze Frame Correlation" approach. The last four columns display the predicted heatmap, where red peaks symbolize the 2D gaze predicted points.

Figure~\ref{fig:qualitative_big_correct} focuses on correct predictions related to \textit{Correct Execution}. The first row shows the actual gaze coordinates. Notably, in the second row corresponding to GLC, the predicted heatmaps exhibit inconsistencies, with varying peaks across consecutive frames. In contrast, our proposed method leverages temporal information to produce temporally consistent predictions. The ``Channel" approach demonstrates better consistency than GLC, while the ``Gaze Frame Correlation" method generates more defined heatmaps with fewer, more localized peaks around the gaze region. In this case, a \textit{Correct Execution} is identified based on the small gap between the ground truth and the predicted gaze trajectory.

Figure~\ref{fig:qualitative_big_mistake} highlights predictions related to \textit{Incorrect Execution}. Here, our approach's gaze predictions diverge from the ground truth, effectively flagging mistakes in action execution.


\end{document}